\documentclass[twoside,11pt]{article}

\usepackage{blindtext}

\usepackage[abbrvbib, preprint]{jmlr2e}
\usepackage{amsfonts,amsmath,amssymb, mathtools}
\usepackage{xcolor}

\renewcommand{\vec}[1]{\boldsymbol{#1}}

\newcommand{\va}{\vec{a}}
\newcommand{\vb}{\vec{b}}

\newcommand{\vg}{\vec{g}}

\newcommand{\vi}{\vec{i}}
\newcommand{\vj}{\vec{j}}
\newcommand{\vk}{\vec{k}}

\newcommand{\vm}{\vec{m}}

\newcommand{\vu}{\vec{u}}
\newcommand{\vv}{\vec{v}}

\newcommand{\vx}{\vec{x}}

\newcommand{\mat}[1]{\mathbf{#1}}

\newcommand{\mA}{\mat{A}}

\newcommand{\mC}{\mat{C}}

\newcommand{\mG}{\mat{G}}

\newcommand{\mI}{\mat{I}}

\newcommand{\mM}{\mat{M}}

\newcommand{\mQ}{\mat{Q}}

\newcommand{\mT}{\mat{T}}
\newcommand{\mU}{\mat{U}}
\newcommand{\mV}{\mat{V}}

\newcommand{\mX}{\mat{X}}

\newcommand{\mZ}{\mat{Z}}
\newcommand{\mPi}{\mat{\Pi}}

\newcommand{\ten}[1]{\boldsymbol{\mathsf{#1}}}

\newcommand{\tA}{\ten{A}}
\newcommand{\tB}{\ten{B}}

\newcommand{\tE}{\ten{E}}

\newcommand{\tG}{\ten{G}}
\newcommand{\tH}{\ten{H}}

\newcommand{\tQ}{\ten{Q}}

\newcommand{\tT}{\ten{T}}
\newcommand{\tU}{\ten{U}}

\newcommand{\tX}{\ten{X}}

\newcommand{\tZ}{\ten{Z}}

\newcommand{\mbC}{\mathbb{C}}

\newcommand{\mbR}{\mathbb{R}}
\newcommand{\mbS}{\mathbb{S}}

\newcommand{\mbV}{\mathbb{V}}

\newcommand{\mcN}{\mathcal{N}}

\newcommand{\mfS}{\mathfrak{S}}

\newcommand{\ii}{{i\mkern1mu}}
\newcommand{\lrp}[1]{\left(#1\right)}
\newcommand{\lrb}[1]{\left[#1\right]}

\newcommand{\tr}[1]{\operatorname{tr}\left(#1\right)}
\newcommand{\expect}[1]{\mathbb{E}\left[#1\right]}
\newcommand{\floor}[1]{\lfloor #1 \rfloor}
\newcommand{\ceil}[1]{\lceil #1 \rceil}
\newcommand{\T}{^\top}
\newcommand{\dx}{\mathrm{d}x}
\newcommand{\dt}{\mathrm{d}t}
\newcommand{\dz}{\mathrm{d}z}
\newcommand{\Mat}{\operatorname{Mat}}
\newcommand{\GOTE}{\mathrm{GOTE}}
\newcommand{\SG}{\mathrm{SG}}

\newcommand{\Sym}{\mathrm{Sym}}
\newcommand{\Sp}{\mathrm{Sp}}

\newcommand{\simiid}{\overset{\mathrm{iid}}{\sim}}
\newcommand{\ssc}{s_{\mathrm{sc}}}
\newcommand{\fsc}{f_{\mathrm{sc}}}

\newcommand{\ind}[1]{\mathbf{1}_{\{#1\}}}
\newcommand{\iid}{i.i.d.}
\newcommand{\aslim}{\xrightarrow[n\to\infty]{\mathrm{a.s.}}}

\definecolor{cw1}{HTML}{b40426}
\definecolor{cw2}{HTML}{e26952}
\definecolor{cw3}{HTML}{f7a889}
\definecolor{cw4}{HTML}{edd1c2}
\definecolor{cw5}{HTML}{c9d7f0}
\definecolor{cw6}{HTML}{9abbff}
\definecolor{cw7}{HTML}{6788ee}
\definecolor{cw8}{HTML}{3b4cc0}

\usepackage{lastpage}

\ShortHeadings{A Random Matrix Analysis of Symmetric Tensor Unfolding}{Szwagier, Mai and Goulart}
\firstpageno{1}

\begin{document}

\title{A Random Matrix Analysis of Symmetric Tensor Unfolding}

\author{\name Tom Szwagier \email tom.szwagier@irit.fr \\
       \addr Univ. Toulouse, Toulouse INP, CNRS, IRIT\\
       31071 Toulouse, France
       \AND
       \name Xiaoyi Mai \email xiaoyi.mai@math.univ-toulouse.fr\\
       \addr Univ. Toulouse, UT2J, IMT\\
       31062 Toulouse, France
       \AND
       \name José Henrique de Morais Goulart \email henrique.goulart@irit.fr\\
       \addr Univ. Toulouse, Toulouse INP, CNRS, IRIT\\
       31071 Toulouse, France}

\editor{My editor}

\maketitle

\begin{abstract}
We study a fundamental problem in machine learning: partial recovery of a planted signal ${\vv \in \mbS^{n-1}}$ from a spiked tensor $\tX = \lambda \, {\vv}^{\otimes p} + \tZ$, where $\tZ$ has symmetric Gaussian entries of variance $\asymp n^{-1}$. While previous works replace $\tZ$ by an \iid\ standard Gaussian tensor to circumvent technical difficulties caused by dependencies, this paper directly analyzes the symmetric model via random matrix theory. We show that the square unfolding algorithm is subject to a sharp phase transition for partial recovery at $\lambda \asymp n^{(p-2) / 4}$, validating an open conjecture from~\citet{montanari_statistical_2014}. We get precise expressions for the limiting spectral distribution, deterministic equivalent, phase transition threshold and overlap. The main message is that unfolded symmetric tensors spectrally behave like standard Wigner matrices, up to simple transformations accounting for high-order symmetries. We consider extensions to multiple orthogonal spikes, recursive unfolding and non-square unfolding via disproportionate random matrix theory, leading to an interesting conjectured statistical--computational tradeoff regarding the choice of unfolding: square versus long.
\end{abstract}


\section{Introduction}

\begin{figure}[t!]
    \centering
    \includegraphics[width=\linewidth]{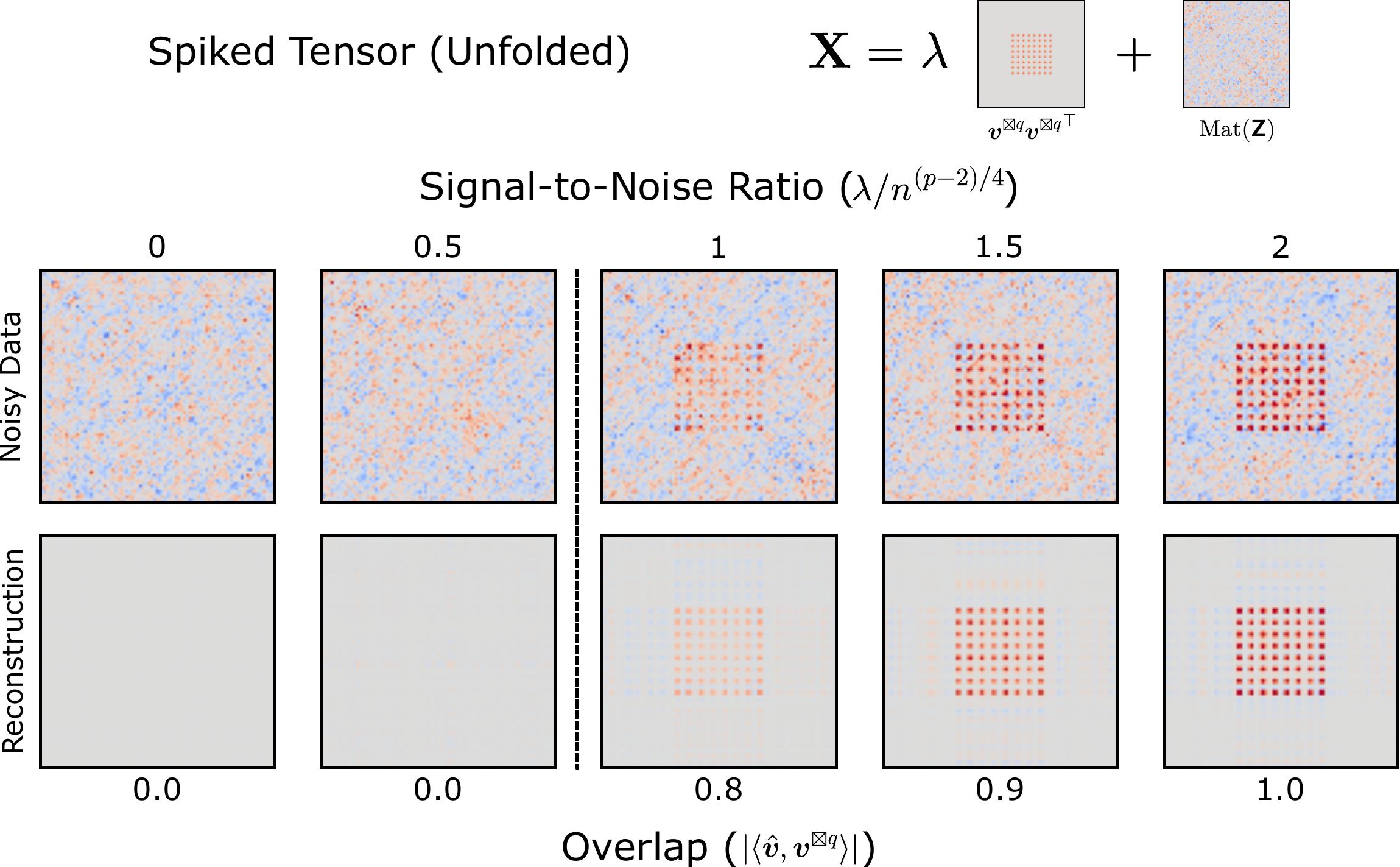}
    \caption{Illustration of the Tensor PCA problem for $p=4$ and $n=20$. Top: spiked tensor model $\tX = \lambda \, {\vv}^{\otimes p} + \tZ$, unfolded into an $n^2 \times n^2$ matrix $\mX$ (see Section~\ref{sec:prelim} for notation). Middle: samples for increasing SNR $\lambda$. Bottom: recovered signal, $\hat{\vv} {\hat{\vv}}\T$, where $\hat{\vv}$ is the leading eigenvector of $\mX$. We observe a phase transition in signal recovery between $0.5$ and $1$, where the overlap goes from 0\% to 80\%.}
    \label{fig:illustration}
\end{figure}

We are interested in the following tensor model:
\begin{equation}\label{eq:spiked_model}
    \tX \coloneqq \lambda \, {\vv}^{\otimes p} + \tZ,
\end{equation}
where $\lambda > 0$ is a \emph{signal-to-noise ratio} (SNR), $\vv \in \mbS^{n-1}$ is a \emph{planted signal} and 
\begin{equation}\label{eq:noise_model}
    \tZ \coloneqq \frac{1}{p!} \sqrt{\frac{p}{n}} \sum_{\sigma\in\mfS_p} \tG^{\sigma}
\end{equation}
is a random tensor of order $p$ and dimension $n$, obtained by symmetrizing a tensor $\tG$ with independent standard Gaussian entries, ensuring rotational invariance.
The goal is to quantify the performance as $n \to \infty$ of a certain estimator of $\vv$ called the \emph{unfolding algorithm}---which amounts to extracting the leading singular vector of a matrix reshaping of the tensor---as a function of the SNR $\lambda$. As usual in this setting, we are also interested in determining the SNR regime under which one can partially recover the signal. An illustration of the problem is given in Figure~\ref{fig:illustration}.

\subsection{Motivation}

Model~\eqref{eq:spiked_model} is a \emph{spiked tensor model}~\citep{montanari_statistical_2014} and the problem of estimating its parameters $\lambda$ and $\vv$ is commonly referred to as \emph{tensor principal component analysis} (Tensor PCA). This generalization of PCA to tensors comes with many applications in machine learning (see reviews from~\citealt{kolda_tensor_2009} and \citealt{sidiropoulos_tensor_2017}). In particular, it is central in the estimation of latent variable models via the method of moments~\citep{pearson_contributions_1894}, with applications to independent component analysis, mixture models, topic models and hidden Markov models~\citep{anandkumar_tensor_2014}. It also arises in community detection, via hypergraph adjacency tensors~\citep{zhou_learning_2006,kim_community_2017,luo_open_2020,luo_tensor_2022}, in collaborative filtering, via tensor completion~\citep{liu_tensor_2013,montanari_spectral_2018} and many image processing tasks going beyond two-dimensional images~\citep{vasilescu_multilinear_2002}. Finally, it serves as a typical model for complex high-dimensional nonconvex landscapes, such as those encountered in deep learning~\citep{janzamin_beating_2016,zhong_recovery_2017,ge_learning_2018,mondelli_connection_2019,arous_landscape_2019,arous_high-dimensional_2024}.

\emph{Unfolding} (sometimes called \emph{matricization}, \emph{reshaping} or \emph{flattening}) is a classical operation in the analysis of tensor-based methods. It is a natural way of transforming a tensor into a matrix, enabling the use of well-established matrix analysis tools, such as eigenvalue decompositions, concentration inequalities~\citep{tropp_introduction_2015} and more general random matrix results~\citep{tao_topics_2012,couillet_random_2022}. Consequently, unfolding is widely used as a spectral relaxation of low-rank tensor decomposition~\citep{romera-paredes_new_2013,mu_square_2014,montanari_statistical_2014}. It is also used as a correlated initialization in many iterative algorithms---such as power iteration~\citep{montanari_statistical_2014,kileel_subspace_2025}, recursive unfolding~\citep{montanari_statistical_2014} and higher-order orthogonal iteration (HOOI)~\citep{de_lathauwer_best_2000}---that achieve exact recovery beyond a critical SNR threshold~\citep{feldman_sharp_2023,lebeau_random_2025}.\footnote{For an estimator $\hat\vv: ({\mbR^n})^{\otimes p} \to \mbS^{n-1}$ of $\vv$, \emph{exact recovery} is here understood as $\liminf_{n\to\infty} |\langle\hat\vv, \vv\rangle| = 1$ and \emph{partial recovery} as $\liminf_{n\to\infty} |\langle\hat\vv, \vv\rangle| > 0$, both almost surely; see~\citet{feldman_sharp_2023}.}

Hence, studying unfolding under spiked tensor models such as~\eqref{eq:spiked_model} is essential for understanding the statistical performance of low-rank tensor decomposition methods, coming with plenty of applications in machine learning. 

\subsection{State of the Art}
Many works have analyzed spiked tensor models, identifying \emph{statistical} and \emph{algorithmic} thresholds on the SNR for signal recovery, under diverse estimation procedures. A statistical threshold refers here to the SNR above which a hypothesis test can distinguish a spiked model $\lambda \, {\vv}^{\otimes p} + \tZ$ from pure noise $\tZ$. An algorithmic threshold refers to the SNR above which a specified computationally-tractable class of estimators can achieve partial recovery. These notions need not coincide and are model- and algorithm-class dependent. The following recovery results are voluntarily simplified for expository purposes.\footnote{We notably omit polylogarithmic factors, as some of them are tightened in follow-up works~\citep{huang_power_2022,wu_sharp_2024,feldman_sharp_2023,bandeira_chaos_2025}.}

Model~\eqref{eq:spiked_model} was originally introduced in~\citet{montanari_statistical_2014}, who proposed several estimators for the planted signal $\vv \in \mbS^{n-1}$, as $n \to \infty$. They showed that maximum likelihood estimation (MLE) achieves partial recovery at $\lambda = \Omega(1)$---that is the statistical threshold~\citep{perry_statistical_2020,jagannath_statistical_2020,goulart_random_2022}. However, solving the MLE problem itself is NP-hard~\citep{hillar_most_2013}, leaving the question of the existence of tractable estimators achieving partial recovery at $\lambda = \Omega(1)$ open.

\begin{figure}[t]
    \centering
    \includegraphics[width=\linewidth]{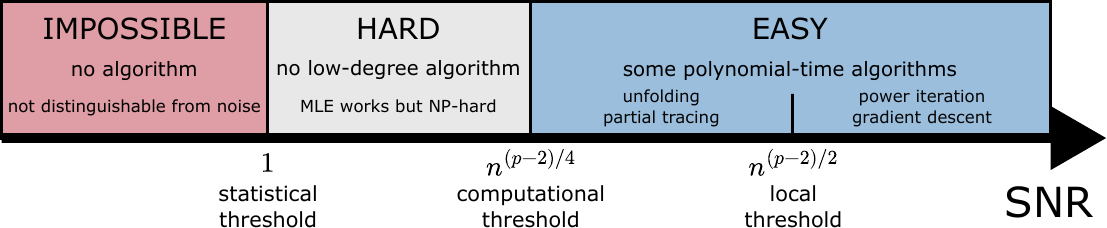}
    \caption{
    Simplified illustration of the conjectured statistical--computational gap for partial recovery, as a function of the SNR $\lambda$. Estimation of the planted signal $\vv \in \mbS^{n-1}$ is impossible below the statistical threshold~\citep{jagannath_statistical_2020}. Above, it is possible by MLE but computationally hard. Below the conjectured computational threshold, no low-degree algorithm achieves recovery~\citep{kunisky_notes_2022}, while above, some polynomial-time algorithms---such as unfolding and partial tracing---do. Above the local threshold, algorithms such as power iteration, approximate message passing and gradient descent work~\citep{montanari_statistical_2014,arous_algorithmic_2020}. Figure adapted from~\citet{wein_mini-course_2024}.
    }
\label{fig:sc_gap}
\end{figure}

Several polynomial-time alternatives for Tensor PCA have then been analyzed. Local methods such as power iteration, approximate message passing, Langevin dynamics and gradient descent have been shown to achieve partial recovery at $\lambda = \Omega(n^{(p-2) / 2})$ under random initialization~\citep{montanari_statistical_2014,arous_algorithmic_2020}. Unfolding, partial tracing and local methods with ``smart'' initializations have been shown to achieve partial recovery at $\lambda = \Omega(n^{(p-2) / 4})$~\citep{montanari_statistical_2014,hopkins_tensor_2015,hopkins_fast_2016,anandkumar_homotopy_2017,arous_long_2023,feldman_sharp_2023}. Several analyses based on sum-of-squares relaxations and low-degree methods suggest that the latter threshold cannot be significantly improved with polynomial-time algorithms~\citep{hopkins_tensor_2015,kunisky_notes_2022}. This leads to a conjectured \emph{statistical--computational} gap, between intractable MLE and tractable algorithms like unfolding that remains open today; see Figure~\ref{fig:sc_gap} for an illustration.

Looking more closely into the assumptions of~\citet{montanari_statistical_2014} regarding tensor unfolding, the main results are actually derived under an \emph{asymmetric} noise model, where the entries of the noise tensor are \emph{independent} standard Gaussians. This importantly differs from the original noise model~\eqref{eq:noise_model}---called \emph{Gaussian Orthogonal Tensor Ensemble} (GOTE, \citealt{gurau_generalization_2020})---whose symmetry yields dependent entries, preventing the application of classical results from random matrix theory. \citet{montanari_statistical_2014} conjecture that the same order of phase transition should hold for the symmetric model.

Some later works partially address this conjecture. Notably, \citet{hopkins_tensor_2015} and \citet{wein_kikuchi_2025} study symmetric models\footnote{The symmetric model considered in~\citet[Problem~3]{hopkins_tensor_2015} is a variant of the original model~\eqref{eq:spiked_model} where the noise entries are \iid\ standard normal up to tensor symmetries. This noise model therefore differs from the GOTE~\eqref{eq:noise_model}. However, it does not change the order of magnitude of the phase transition, as we will discuss ahead (see Remark~\ref{rem:standard}). \label{foot:standard}} and obtain $\Omega(n^{(p-2) / 4})$ recovery guarantees under variants of the tensor unfolding algorithm, while not providing either sharp transitions or precise overlaps. Follow-up works obtain sharp unfolding transitions and overlaps for even and odd orders~\citep{arous_long_2023,feldman_sharp_2023,lebeau_random_2025}, but all under an asymmetric noise model that is not adapted to symmetric signals.

Hence, a random matrix analysis of the unfolding algorithm on the genuinely symmetric spiked tensor model of~\citet{montanari_statistical_2014} is currently lacking in the Tensor PCA literature.

\subsection{Contributions}\label{subsec:contrib}

We perform a random matrix analysis of the unfolding algorithm under the spiked tensor model~\eqref{eq:spiked_model}, with \emph{even} order ($p \coloneqq 2q$) and \emph{square} unfolding ($n^q \times n^q$).
Specifically, we derive a {deterministic equivalent} \citep{couillet_random_2022} of the resolvent and deduce the weak convergence of the {empirical spectral distribution} towards a generalization of the Wigner semicircle distribution~\citep{wigner_distribution_1958}, that is a mixture of a  \emph{scaled semicircle} distribution and a \emph{Dirac} distribution, whose weights depend on the tensor order $p$. Despite the non-classical structure of the resolvent and the limiting spectral distribution, we show that the unfolding algorithm undergoes a classical phase transition regarding the recovery of the \emph{lifted} signal $\vv^{\boxtimes q} \in \mbS^{n^q-1}$, with explicit critical value and overlap---$\boxtimes$ denotes the Kronecker product. See Section~\ref{sec:prelim} for notation and definitions.

More precisely, let $\tZ$ be a Gaussian orthogonal tensor defined as in~\eqref{eq:noise_model} and $\mZ$ its $n^q \times n^q$ unfolding. Denoting by $\fsc(x) \coloneqq \sqrt{(4 - x^2)^+} / (2 \pi)$ the standard Wigner semicircle density and $\alpha_{p, q} \coloneqq \sqrt{q! / (p - 1)!}$, the empirical spectral measure of $\alpha_{p, q}^{-1} \, n^{-(p-2) / 4} \, \mZ$ converges weakly almost surely to $q!^{-1} \, \fsc(x) \,  \dx + (1 - q!^{-1}) \, \delta_0(\dx)$. Let $\tX = \lambda \, {\vv}^{\otimes p} + \tZ$ be a spiked tensor as in~\eqref{eq:spiked_model}, $\mX$ its $n^q \times n^q$ unfolding and $(\hat\lambda, \hat \vv) \in \mbR_+\times\mbS^{n^q-1}$ its leading eigenpair. Writing $\tau \coloneqq \alpha_{p, q}^{-1} \, n^{-(p-2) / 4} \, \lambda$ and $\hat\tau \coloneqq \alpha_{p, q}^{-1} \, n^{-(p-2) / 4} \, \hat\lambda$, one has, almost surely, as $n\to\infty$, the following sharp phase transitions: for $\tau \leq 1$, $\hat\tau \to 2$ and $|{\hat \vv}\T \vv^{\boxtimes q}|^2 \to 0$ while for $\tau > 1$,  $\hat\tau \to \tau + \tau^{-1}$ and $|{\hat \vv}\T \vv^{\boxtimes q}|^2 \to 1 - \tau^{-2}$.

Our results confirm the open conjectures from~\citet{montanari_statistical_2014} and \citet{hopkins_tensor_2015} on symmetric spiked tensor unfolding: there is indeed a phase transition for partial recovery at $\lambda \asymp n^{(p-2) / 4}$ and we now know its precise constant: $\alpha_{p, q} \coloneqq \sqrt{(p / 2)! / (p - 1)!}$. Hence, the high-order symmetries do not affect significantly the order of magnitude of the phase transition for partial recovery with respect to asymmetric models. They however significantly affect the spectral distribution, which becomes a mixture of a semicircle and a Dirac distribution. Moreover, although the phase transition is at the conjectured order of magnitude, $\lambda \asymp n^{(p-2) / 4}$, it appears at a different scale from the asymmetric model and the effect increases with the tensor order according to $\alpha_{p, q}$.\footnote{Whether the phase transition under the symmetric model appears ``before'' or ``after'' the phase transition under the asymmetric model depends on the choice of symmetric model. Under GOTE noise, it appears before, according to $\alpha_{p, q} \coloneqq \sqrt{(p / 2)! / (p - 1)!} \leq 1$ (see Theorem~\ref{thm:transition}), while under \iid\ standard Gaussian noise (up to tensor symmetries), it appears after, according to $\beta \coloneqq \sqrt{(p / 2)!} \geq 1$ (see Remark~\ref{rem:standard}).}
Therefore, a simplified analysis focused on a ``symmetric signal + asymmetric noise'' model~\citep{montanari_statistical_2014, kim_community_2017, feldman_sharp_2023} does not lead to the same phase transition thresholds and overlaps obtained under the genuine spiked tensor model~\eqref{eq:spiked_model} and misses important spectral phenomena caused by the loss of independence induced by the symmetries.

The proof relies on \emph{random matrix theory} (RMT), exposing important phenomena that cannot be revealed via classical machine learning tools---notably, a sharp phase transition in spectral behavior (the so-called \emph{BBP transition} of~\citealt{baik_phase_2005}), with precise eigenvector overlaps. The high-order symmetries in the GOTE break the fundamental independence assumption required in standard random matrix theory. To alleviate these dependencies, we project the unfolded GOTE onto symmetric tensor subspaces, which boils down to removing duplicated rows and columns and applying a specific scaling on the entries. The whole procedure, that we name \emph{symmetric unfolding}, preserves the eigenstructure (up to a simple isometry), while removing the most impacting dependencies. The resulting matrix spectrally behaves like a standard Wigner matrix, from which we can precisely characterize the limiting spectral distribution and overlaps.

The paper is organized as follows. 
In Section~\ref{sec:prelim}, we introduce our notation, the spiked tensor model, the unfolding map and the main random matrix tools needed to establish our results. 
In Section~\ref{sec:results}, we state the main results of the paper: the analysis of unfolded symmetric tensors under the framework of hypersymmetric matrices, the deterministic equivalent, the limiting spectral distribution and the eigenvector overlap. We run simulations and show a strong match with theory, even for relatively small dimensions, $n \in \{20, 50, 100\}$. 
In Section~\ref{sec:discussion}, we discuss and put the previous results into perspective.
In Appendix~\ref{sec:proofs}, we give the proofs of the main results. 
In Appendix~\ref{sec:standard}, we extend our results and simulations to the \iid\ symmetric standard Gaussian tensor model.
In Appendix~\ref{sec:multispike}, we extend our results to multiple spikes, under the canonical polyadic (CP) model.
In Appendix~\ref{sec:recursive}, we analyze a recursive unfolding procedure to estimate the planted signal $\vv \in \mbS^{n-1}$ from the lifted signal $\vv^{\boxtimes q} \in \mbS^{n^q-1}$; we conjecture a sharp phase transition at $\lambda = \alpha_{p, q} \, n^{(p-2) / 4}$, from null to exact recovery.
In Appendix~\ref{sec:choice_q}, we study odd order $p$ and non-square unfolding ($n^k \times n^{p-k}$, with $1 \leq k < p/2$); we conjecture a sharp phase transition at $\lambda = \gamma_{p, k} \, n^{(p-2)/4}$, with $\gamma_{p, k} \coloneqq (k!(p-k)! / (p-1)!^2)^{1/4}$, resulting in interesting statistical--computational tradeoffs.

\section{Preliminaries}\label{sec:prelim}
\subsection{Notation}
Throughout this paper, we adopt the following notation. The dimension of the tensor is denoted by $n$ and assumed to be large, while the order is denoted by $p$ and assumed to be fixed and even: $p \coloneqq 2 q$ for some (fixed) positive integer $q$. For convenience, integer intervals are shortened as $[n] \coloneqq \{1, \ldots, n\}$ and multi-indices as $\vi \coloneqq (i_1, \ldots, i_p) \in \lrb{n}^p$. The set of permutations of a multi-index $\{i_1, \ldots, i_p\}$ is written $\mfS_{i_1, \ldots, i_p}$; for short, we write $\mfS_p \coloneqq \mfS_{1, \ldots, p}$. The cardinal of a set is denoted by $\#$. We also denote by $\vi \sqcup \vj$ the \emph{multiset} union of two multi-indices $\vi$ and $\vj$. Scalars, vectors, matrices and tensors are respectively written with the following fonts: $a$, $\va$, $\mA$ and $\tA$. We follow the common convention of omitting the indices in sequences of (random or deterministic) matrices: $\va \in \mbR^n$ and $\mA \in \mbR^{n \times n}$ should be respectively understood as $\va_n \in \mbR^n$ and $\mA_n \in \mbR^{n \times n}$.

Outer and Kronecker products of vectors $\va_1, \ldots, \va_p \in \mbR^n$ are respectively denoted by $\va_1 \otimes \cdots \otimes \va_p \in ({\mbR^n})^{\otimes p}$ and $\va_1 \boxtimes \cdots \boxtimes \va_p \in \mbR^{n^p}$; in addition, we write $\va^{\otimes p} = \va \otimes \cdots \otimes \va$ and $\va^{\boxtimes p} = \va \boxtimes \cdots \boxtimes \va$, where the product is repeated $p$ times. The spectrum of a symmetric matrix $\mM \in \mbR^{n \times n}$ is denoted by $\Sp(\mM)$; its eigenvalues, $\lambda_1(\mM) \geq \cdots \geq \lambda_n (\mM)$, are repeated according to their multiplicity and sorted in decreasing order. 

As we work in the asymptotic regime where $n \to \infty$, we use the classical \emph{Bachmann--Landau} ``big $O$'' notation. We write $f(n) = O(g(n))$ if there exists a positive constant $C$ such that $|f(n)| \leq C \, g(n)$ as $n\to\infty$. Similarly, we write $f(n) = \Omega(g(n))$ if $g(n) = O(f(n))$, $f(n) \asymp g(n)$ if $f(n) = O(g(n))$ and $g(n) = O(f(n))$. Finally, we write $f(n) = o(g(n))$ if $f(n) / g(n) \to 0$, $f(n) = \omega(g(n))$ if $f(n) / g(n) \to \infty$ and $f(n) \sim g(n)$ if $f(n) / g(n) \to 1$.

\subsection{Symmetric Tensors}\label{subsec:sym}
The spiked tensor model~\eqref{eq:spiked_model} we consider in this work has the following entries:
\begin{equation*}
    {X}_{i_1 \cdots i_p} = \lambda \, {v}_{i_1} \cdots {v}_{i_p} + \frac{1}{p!} \sqrt{\frac{p}{n}} \sum_{\sigma\in\mfS_p} {G}_{i_{\sigma(1)}  \cdots i_{\sigma(p)}},
\end{equation*}
with $\lambda > 0$, $\vv \in \mbS^{n-1}$ and ${G}_{i_1 \cdots i_p} \simiid \mcN(0, 1)$. The random tensor $\tZ \coloneqq \frac{1}{p!} \sqrt{\frac{p}{n}} \sum_{\sigma\in\mfS_p} \tG^{\sigma}$ belongs to the Gaussian Orthogonal Tensor Ensemble; we write $\tZ \sim \GOTE_{p, n}$.\footnote{Different definitions for the GOTE exist in the literature~\citep{gurau_generalization_2020,mukherjee_spectra_2025,bonnin_characterization_2026,bonnin_universality_2026}. They only differ by the exponent they put on the dimension $n$, which depends on the estimator under analysis. In our case, following the original work of~\citet{montanari_statistical_2014}, we have entries with variance $\asymp n^{-1}$.} 

One can show that the entries of $\tZ$ are independent normal---\emph{up to tensor symmetries}---and their variance depends on the multiplicities of the multi-index~\citep{bonnin_characterization_2026}. More precisely, for $i_1 \leq \cdots \leq i_p$ and $\sigma \in \mfS_p$, one has
\begin{equation*}
    \begin{cases}
        {Z}_{i_1 \cdots i_p} &\simiid \mcN\lrp{0, \frac{p}{n \#\mfS_{i_1, \ldots, i_p}}},\\
        {Z}_{i_{\sigma(1)}  \cdots i_{\sigma(p)}} &= {Z}_{i_1 \cdots i_p}.
    \end{cases}
\end{equation*}
Hence, the GOTE is a generalization of the (matrix) \emph{Gaussian Orthogonal Ensemble} (GOE), which exhibits variance $1 / n$ off-diagonal and $2/n$ on-diagonal, to $p \geq 2$.

From a combinatorial viewpoint, writing $\vm$ the multiset of multiplicities of the indices in $\vi \in [n]^p$, one has 
\begin{equation}
    \#\mfS_{\vi} = \binom{p}{\vm} \coloneqq \frac{p!}{\prod_{m \in \vm} (m!)}.
\end{equation}
The latter expression is called a \emph{multinomial coefficient}---a multi-way generalization of a binomial coefficient.
The set of $p$-indices with multiplicities $\vm$ therefore has cardinal $\binom{n}{\#\vm}\binom{p}{\vm}$. This implies that multi-indices with distinct indices predominate in high dimensions; indeed, whenever $\vm \neq \{1, \ldots, 1\}$, one has $\#\vm < p$ therefore $\binom{n}{\#\vm} / \binom{n}{p} \to 0$.

Let $\Sym^p(\mbR^n)$ denote the space of symmetric tensors of order $p$ and dimension $n$---we remove the superscript when $p=2$ for symmetric matrices. Then $\Sym^p(\mbR^n)$ is a linear subspace of $({\mbR^n})^{\otimes p}$ of dimension $n_p \coloneqq \dim(\Sym^p(\mbR^n)) = \binom{n+p-1}{p}$; this can be established via standard ``stars-and-bars'' arguments. A canonical orthonormal basis of $\Sym^p(\mbR^n)$ is
\begin{equation}\label{eq:basis}
    \Biggl\{\tU_{\vi} \coloneqq  \frac{1}{{\sqrt{\#\mfS_{\vi}}}} \sum_{\vj \in \mfS_{\vi}} \tE_{\vj} \colon\quad 1 \leq i_1 \leq \cdots \leq i_p \leq n\Biggr\},
\end{equation}
where $\tE_{\vj} \in ({\mbR^n})^{\otimes p}$ is the unit tensor with zeros everywhere except in position $(j_1, \ldots, j_p)$.

\subsection{Tensor Unfolding}\label{subsec:unfolding}
\label{sec:unfold}
Let us define the following bijective function between multi-indices and indices:
\begin{equation}
    \begin{split}
        \varphi_k\colon {[n]}^k &\longrightarrow [n^k]\\
        (i_1, \ldots, i_k) &\longmapsto 1 + \sum_{j=1}^k (i_j - 1) \, n^{j-1}.
    \end{split}
\end{equation}
The map $\varphi_k$ induces a natural bijection between order-$p$ tensors and matrices that is commonly called \emph{unfolding}:
\begin{equation}
    \begin{split}
        \Mat_k\colon ({\mbR^n})^{\otimes p} &\longrightarrow \mbR^{n^k \times n^{p-k}}\\
        \lrp{{X}_{i_1 \cdots i_p}}_{(i_1, \ldots, i_p) \in \lrb{n}^p} &\longmapsto \lrp{{X}_{\varphi_k^{-1}(i) ; \varphi_{p-k}^{-1}(j)}}_{(i, j) \in [n^q] \times [n^{p-k}]}.
    \end{split}
\end{equation}
We identify $\Mat_p$ with the classical vectorization $\operatorname{vec}\colon ({\mbR^n})^{\otimes p} \to \mbR^{n^p}$. In this paper, we choose $k = \floor{p/2} = q$. This corresponds to a \emph{square unfolding}, yielding an $n^q \times n^q$ symmetric matrix---like a classical Wigner matrix but with extra symmetries. 
\begin{remark}[Choice of Unfolding]\label{rem:unfolding}
    Other choices of $k$ for unfolding would have led to so-called \emph{long}~\citep{arous_long_2023} or \emph{disproportionate}~\citep{feldman_spiked_2023} random matrices. In that case, the dimensions ($n^k \times n^{p-k}$) are no longer comparable; one cannot apply classical RMT results~\citep{benaych-georges_singular_2012} and needs to use more sophisticated tools~\citep{alex_isotropic_2014}.
    In addition, the choice of an unfolding that is ``as square as possible'' is recommended in~\citet{montanari_statistical_2014} and \citet{mu_square_2014}, as it minimizes the operator norm of the unfolded noise. It is worth noting that this heuristic is later nuanced by~\citet{arous_long_2023}, which states that the SNR threshold does not depend on the choice of $k$ in the \emph{asymmetric} model and therefore recommend $k=1$ to avoid an extra recursive unfolding step to recover the planted signal. We shed light on these opposite points of view in Section~\ref{sec:discussion} and Appendix~\ref{sec:choice_q}.
\end{remark}
\noindent With a slight abuse of notation, we will often replace a multi-index $\vi\coloneqq(i_1, \ldots, i_k)\in{[n]}^k$ by its associated index $i\coloneqq \varphi_k(\vi)\in[n^k]$ and vice-versa. Likewise, we will often assimilate a tensor $\tX \in ({\mbR^n})^{\otimes p}$ to its associated matrix $\mX \coloneqq \Mat_k(\tX) \in \mbR^{n^k \times n^{p-k}}$.

\subsection{Useful Tools from Random Matrix Theory}\label{subsec:RMT}

Random matrix theory is a rich mathematical field at the intersection of probability theory, linear algebra and complex analysis. It provides tools to study the eigenstructure of large matrices with random entries---notably their limiting spectral distribution, isolated eigenvalues and associated eigenspaces---enabling the theoretical performance analysis of many machine learning algorithms. The RMT framework offers several advantages over more classical theoretical tools in machine learning, often referred to as the \emph{blessing of dimensionality}~\citep{donoho_high-dimensional_2000}. In high dimensions, a priori random spectral quantities often become asymptotically deterministic through concentration of measure, while universality makes their limiting behavior largely insensitive to the precise noise distribution. Moreover, RMT accommodates structured and dependent noise, as well as certain nonlinear spectral statistics that are central in machine learning (kernel, activation function). These features make it particularly well suited to characterizing phase transitions and partial recovery in high-dimensional planted-signal models. This subsection recalls the main results from RMT that will be used throughout this paper. Its content is mostly based on~\citet{couillet_random_2022}, where the reader can find more details.

The first object we review is the \emph{resolvent}, which is central in RMT, as it provides information on the eigenstructure.
\begin{lemma}[Eigenstructure via Resolvent]\label{lem:resolvent}
   Let $\mM \in \mbR^{n \times n}$ be a symmetric matrix, with spectrum $\Sp(\mM) \in \mbR^n$.
   Let $\mu_{\mM} \coloneqq \frac{1}{n} \sum_{i=1}^n \delta_{\lambda_i(\mM)}$ be its \emph{empirical spectral measure}, $s_{\mu_{\mM}}(z) \coloneqq \int 1 / (t-z) \, \mu_\mM(\dt)$ its \emph{Stieltjes transform} and $\mQ_{\mM}(z) \coloneqq (\mM - z \, \mI_n)^{-1}$ its \emph{resolvent} (with $z \in \mbC \setminus\Sp(\mM)$). Let $\lambda \in \Sp(\mM)$ be a \emph{simple} eigenvalue of $\mM$, with associated eigenvector $\hat\vu\in \mbS^{n-1}$ and $\Gamma$ a \emph{closed contour} around $\lambda$ \emph{only}. Then, one has 
   \begin{equation}
       s_{\mu_\mM}(z) = \frac{1}{n} \tr{\mQ_\mM(z)} \quad \text{and  } \quad |{\hat\vu}\T \vu|^2 = - \frac{1}{2 \pi \ii} \oint_{\Gamma} \vu\T \mQ_\mM(z) \vu \, \dz \quad (\vu \in \mbS^{n-1}).
   \end{equation}
\end{lemma}
Hence, the resolvent gives access to the empirical spectral measure and the eigenvector overlap of \emph{deterministic} matrices.\footnote{
The spectral measure can be recovered via the Stieltjes inversion formula: if $a, b \in \mbR$ are continuity points of a probability measure $\mu$, then $\mu([a, b]) = \lim_{y \downarrow 0} \int_a^b \Im[s_\mu(x + \ii y)] / \pi \, \dx$, while if $\mu$ has an isolated mass at $x \in \mbR$, then $\mu(\{x\}) = \lim_{y \downarrow 0} - \ii y s_\mu(x + \ii y)$. Moreover, the pointwise convergence of $s_\mu$ implies the weak convergence of $\mu$ due to the \emph{Stieltjes continuity theorem} (see, e.g., Exercise~2.4.10 of~\citealt{tao_topics_2012}).} When these are \emph{random}, the eigenstructure can be characterized via \emph{deterministic equivalents}~\citep{girko_circular_1985,hachem_deterministic_2007}, i.e. deterministic matrices that ``behave'' similarly asymptotically with respect to certain linear and bilinear forms, as defined next.
\begin{definition}[Deterministic Equivalent]\label{def:DE}
    Let $\mQ \in \mbR^{n\times n}$ be a random matrix. We say that $\bar \mQ \in \mbR^{n\times n}$ is a \emph{deterministic equivalent} of $\mQ$---and write $\mQ \leftrightarrow \bar \mQ$---if, for any sequences of deterministic matrices $\mA \in \mbR^{n \times n}$ of unit operator norm and vectors $\va, \vb \in \mbR^n$ of unit Euclidean norm, one has, almost surely, as $n \to \infty$,
    \begin{equation}
       \label{eq:det_equiv}
        \frac{1}{n} \tr{\mA \lrp{\mQ - \bar\mQ}} \aslim 0 \qquad \text{and} \qquad \va\T \lrp{\mQ - \bar\mQ} \vb \aslim 0.
    \end{equation}
\end{definition}
Such deterministic equivalents can be obtained by showing the concentration of the expressions in~\eqref{eq:det_equiv} around their expectation. To that end, we state two fundamental probabilistic identities, where $\expect{\cdot}$ and $\mbV[\cdot]$ denote respectively the expectation and the variance.
\begin{lemma}[Stein Lemma, \citealt{stein_estimation_1981}]\label{lem:stein}
    Let $x \sim \mcN(0, 1)$ and $f\colon \mbR \to \mbR$. Then, upon existence of the following quantities, one has 
    \begin{equation}
        \expect{x \, f(x)} = \expect{f'(x)}.
    \end{equation}
\end{lemma}
\begin{lemma}[Nash--Poincaré Inequality]\label{lem:np}
    Let $\vx \sim \mcN(0, \mC)$, with $\mC \in \mbR^{n \times n}$ and $f\colon \mbR^n \to \mbR$. Then, upon existence of the following quantities, one has 
    \begin{equation}
        \mbV\lrb{f(\vx)} \leq \sum_{i, j \in [n]} \mC_{ij} \expect{\frac{\partial f(\vx)}{\partial{x}_i} \frac{\partial f(\vx)}{\partial{x}_j}}.
    \end{equation}
\end{lemma}
Now that the main tools are given, let us outline a typical approach to characterize eigenstructure via RMT. The first step is to establish a deterministic equivalent for the resolvent matrix $\mQ_\mM \in \mbR^{n \times n}$, a natural candidate being its expectation, $\expect{\mQ_\mM}$. To achieve this, we derive a self-consistent equation for the resolvent and simplify it using Stein’s Lemma (Lemma~\ref{lem:stein}). This yields an explicit asymptotic expression for the expected resolvent. Next, we demonstrate the concentration of linear and bilinear forms of the resolvent around their expectation via the Nash--Poincaré inequality (Lemma~\ref{lem:np}). If the resulting variance bound drops fast enough, then this allows establishing almost sure deterministic equivalence (Definition~\ref{def:DE}). The latter yields the \emph{limiting spectral distribution} via inverse Stieltjes transform and the \emph{eigenvector overlap} $|{\hat \vu}\T \vu|$ via Cauchy’s integral formula (see Lemma~\ref{lem:resolvent}). Finally, the integral expression can be further simplified using the Sherman--Morrison formula, which relates the resolvent of the spiked model to that of the noise model. We specialize this scheme to the symmetric unfolding in Appendix~\ref{sec:proofs}.

\section{Main Results}\label{sec:results}
This section states the main results, namely the impact of tensor symmetries on the spectrum, the deterministic equivalent of the resolvent---consequently, the limiting spectral distribution---and the spiked eigenvalues and overlaps. The proofs are given in Appendix~\ref{sec:proofs}.

\subsection{Hypersymmetric Matrices and Symmetric Unfolding}\label{subsec:hypersym}
Let $\tZ \sim \GOTE_{p, n}$ and $\mZ \coloneqq \Mat_q(\tZ) \in \mbR^{n^q \times n^q}$. As $\mZ$ is a symmetric matrix with (mostly) independent Gaussian entries, one could think that its eigenvalues are almost-surely simple and nonzero---by applying Sard's theorem to the discriminant polynomial~\citep{breiding_geometry_2018}. However, the symmetries in $\tZ$ lead its unfolding $\mZ$ to have \emph{hypersymmetries}---i.e., symmetries beyond the classical matrix ones---including notably duplicated columns. The next definition formalizes this notion of hypersymmetry in symmetric tensors. 
\begin{definition}[Hypersymmetric Matrix]
    We say that $\mM \in \mbR^{n^q \times n^q}$ is a \emph{hypersymmetric matrix} if there exists $\tT \in \Sym^p(\mbR^n)$ such that $\mM = \Mat_q(\tT)$. Equivalently,
    \begin{equation}
        {M}_{ij} = {M}_{i'j'}, \qquad \forall 
        (i_1', \ldots, i_q', j_1', \ldots, j_q') \in \mfS_{i_1, \ldots, i_q, j_1, \ldots, j_q}.
    \end{equation}
\end{definition}
An important consequence of hypersymmetry is \emph{duplicated rows and columns}: for $(\vi', \vj') \in \mfS_{\vi} \times \mfS_{\vj}$, one has ${M}_{i \cdot} = {M}_{i' \cdot}$ and ${M}_{\cdot j} = {M}_{\cdot j'}$. Intuitively, removing these duplicates does not change the eigenstructure. We formalize this observation by projecting the row and column spaces onto symmetric tensor subspaces.

Let $\mathcal I_q \coloneqq \left\{(i_1,\ldots,i_q)\in[n]^q\colon i_1\leq\cdots\leq i_q\right\}$ and $n_q \coloneqq |\mathcal I_q| = \binom{q+n-1}{q}$. For $\vi \in \mathcal I_q$, define $\vu_{\vi} \coloneqq \operatorname{vec}(\tU_{\vi}) \in \mbS^{n^q - 1}$~\eqref{eq:basis} and $\mU \coloneqq \lrp{\vu_{\vi}}_{\vi \in \mathcal I_q} \in \mbR^{n^q \times n_q}$. The vectors $(\vu_{\vi})_{\vi\in\mathcal I_q}$ form an \emph{orthonormal basis} of $\Sym^q(\mbR^n)$; hence $\mU\T\mU=\mI_{n_q}$ and $\mPi \coloneqq \mU \mU\T = {(\#\mfS_{\vi}^{-1} \ind{\vj \in \mfS_{\vi}})}_{i,j \in [n^q]}$ is the \emph{orthogonal projector} onto $\Sym^q(\mbR^n)$.
\begin{proposition}[Eigenstructure of Hypersymmetric Matrices]\label{prop:eigenstructure}
    Let $\tT \in \Sym^p(\mbR^n)$ and $\mM \coloneqq \Mat_q(\tT) \in \Sym(\mbR^{n^q})$. Then, one has
    \begin{equation}
        \mM = \mU \mM' \mU\T, \qquad \text{with} \qquad \mM' \coloneqq \mU\T \mM \mU = \bigl(\sqrt{\#\mfS_{\vi} \#\mfS_{\vj}} \, \tT_{\vi\vj}\bigr)_{\vi, \vj \in \mathcal I_q}.
    \end{equation}
    Consequently, one has: 
    \begin{enumerate}
        \item $\Sp(\mM) = \Sp(\mM') \cup \{0\}, \qquad$ with $\dim(\ker(\mM)) \geq n^q - \binom{q + n - 1}{q} \sim \bigl(1 - \frac{1}{q!}\bigr) \, n^q $.
        \item $\forall (\lambda', \vv') \in \mbR\times\mbS^{n_q - 1}, \quad \mM' \vv' = \lambda' \vv' \Longleftrightarrow \mM \mU\vv' = \lambda' \mU\vv'$.
    \end{enumerate}
\end{proposition}
We call $\Phi_q\colon \tT \mapsto \mU\T \Mat_q(\tT) \mU$ the \emph{symmetric unfolding}. The previous result implies that the nonzero eigenpairs of $\Mat_q(\tT)$ and $\Phi_q(\tT)$ are isometric; we may therefore abusively switch between the two representations in the following.

Although elementary, Proposition~\ref{prop:eigenstructure} is actually extremely informative: the eigenstructure of $\Mat_q(\tT)$---notably the limiting spectral distribution, the spiked eigenvalues and overlaps---can be simply obtained via $\Phi_q(\tT)$, a compact version obtained by removing duplicated columns and scaling by the number of permutations of the indices.\footnote{Note that the overlaps in $\mbS^{n^q - 1}$ and $\mbS^{n_q - 1}$ are \emph{exactly} equal, since $\mU$ is an isometry.} From a practical standpoint, this yields a computational speedup by a factor $(n^q / n_q)^3 \sim q!^3$ for eigendecomposition. From a theoretical standpoint, the projection onto the symmetric subspace reduces the number of dependencies between the entries. Let 
\begin{equation}\label{eq:equivalence}
    \mathcal E(\vi,\vj) \coloneqq \left\{(\vi',\vj')\in\mathcal I_q^2\colon \vi'\sqcup\vj'=\vi\sqcup\vj\right\}    
\end{equation}
represent the equivalence class of entries that are determined by the entry $(\vi,\vj) \in \mathcal I_q^2$. Then $\#\mathcal E(\vi,\vj) \leq \binom{p}{q} = p! / q!^2$ and we later show that these extra dependencies do not impact the asymptotic eigenstructure; see Section~\ref{sec:discussion} for a broader discussion.
\begin{example}[$p = 4$, $n=2$]
    Let $\tT \coloneqq (T_{ijkl})_{i, j, k, l \in \{1, 2\}} \in \Sym^4(\mbR^2)$. Then,
    \begin{equation*}
        \Mat_q(\tT) = 
        \begin{bmatrix}
            \textcolor{cw8}{t_{1111}} & \textcolor{cw6}{t_{1121}} & \textcolor{cw6}{t_{1112}} & \textcolor{gray}{t_{1122}}\\
            \textcolor{cw6}{t_{2111}} & \textcolor{gray}{t_{2121}} & \textcolor{gray}{t_{2112}} & \textcolor{cw3}{t_{2122}}\\
            \textcolor{cw6}{t_{1211}} & \textcolor{gray}{t_{1221}} & \textcolor{gray}{t_{1212}} & \textcolor{cw3}{t_{1222}}\\
            \textcolor{gray}{t_{2211}} & \textcolor{cw3}{t_{2221}} & \textcolor{cw3}{t_{2212}} & \textcolor{cw1}{t_{2222}}
        \end{bmatrix}
        \text{ and }
        \Phi_q(\tT) = 
        \begin{bmatrix}
            \textcolor{cw8}{1 \, t_{1111}} & \textcolor{cw6}{\sqrt{2} \, t_{1121}} & \textcolor{gray}{1 \, t_{1122}}\\
            \textcolor{cw6}{\sqrt{2} \, t_{2111}} & \textcolor{gray}{2 \, t_{2121}} & \textcolor{cw3}{\sqrt{2} \, t_{2122}}\\
            \textcolor{gray}{1 \, t_{2211}} & \textcolor{cw3}{\sqrt{2} \, t_{2221}} & \textcolor{cw1}{1 \, t_{2222}}
        \end{bmatrix}.
    \end{equation*}
    The colors depict the hypersymmetries.
\end{example}
\begin{remark}[Catalecticant Matrices]
    The symmetric unfolding can be identified, up to the normalization induced by the orthonormal basis, with \emph{catalecticant matrices}, whose determinant is the catalecticant polynomial from invariant theory. This connection links the present spectral approach to the classical catalecticant methods of Sylvester and their modern extensions~\citep{sylvester_remarkable_1851,brachat_symmetric_2010}. In the noiseless setting, these methods exploit the low-rank and kernel structure of the catalecticant to recover symmetric rank decompositions. This suggests a hybrid approach where spectral denoising is followed by catalecticant methods for recovery. We leave this perspective for future research.
\end{remark}

\subsection{Deterministic Equivalent}\label{subsec:DE}
The tools presented in previous sections (\ref{subsec:RMT} and \ref{subsec:hypersym}) to analyze random matrices and hypersymmetric ones can now be combined to characterize the spectrum of square unfoldings of symmetric tensors under the spiked model~\eqref{eq:spiked_model}.
The following theorem gives a deterministic equivalent of the noise resolvent.
\begin{theorem}[Deterministic Equivalent]\label{thm:DE_symmetric}
    Let $\tZ \sim \GOTE_{p, n}$, $\mM \coloneqq \alpha_{p, q}^{-1} \, n^{-(p-2) / 4} \, \Phi_q(\tZ)$ and $\mQ(z) \coloneqq \lrp{\mM - z \, \mI_{n_q}}^{-1}$, with $\alpha_{p, q} \coloneqq \sqrt{q! / (p - 1)!}$. Then,
    \begin{equation}\label{eq:resolvent_symmetric}
        \mQ(z) \longleftrightarrow \ssc(z) \, \mI_{n_q}, \qquad \text{with} \qquad \ssc(z) \coloneqq - \bigl(z - \sqrt{z^2 - 4}\bigr) / 2,
    \end{equation}
   in the sense of Definition~\ref{def:DE}. Consequently, the empirical spectral distribution of $\mM$ converges weakly almost surely to the standard Wigner semicircle law:
    \begin{equation}\label{eq:lsd_symmetric}
        \frac{1}{n_q}\sum_{i=1}^{n_q}\delta_{\lambda_i(\mM)} \aslim \fsc(x)\,\dx, \qquad \text{with} \qquad  \fsc(x) = \frac{\sqrt{(4-x^2)^+}}{2\pi}.
    \end{equation}
\end{theorem}
The theorem says that, on the symmetric tensor subspace, the normalized GOTE unfolding has the same deterministic resolvent as a classical GOE matrix. As a consequence of Proposition~\ref{prop:eigenstructure}, we get a deterministic equivalent of the GOTE unfolding.
\begin{corollary}[Standard Unfolding]\label{cor:DE}
    Let $\tZ \sim \GOTE_{p, n}$, $\mM \coloneqq \alpha_{p, q}^{-1} \, n^{-(p-2) / 4} \, \Mat_q(\tZ)$ and $\mQ(z) \coloneqq \lrp{\mM - z \, \mI_{n^q}}^{-1}$, with $\alpha_{p, q} \coloneqq \sqrt{q! / (p - 1)!}$. Then,
    \begin{equation}\label{eq:resolvent}
        \mQ(z) \longleftrightarrow \ssc(z) \, \mPi + s_{\delta_0}(z) \, (\mI_{n^q} - \mPi), \qquad \text{with} \qquad s_{\delta_0}(z) \coloneqq -\frac{1}{z},
    \end{equation}
    in the sense of Definition~\ref{def:DE}. Consequently, the empirical spectral distribution of $\mM$ converges weakly almost surely to a mixture of a semicircle and a Dirac:
    \begin{equation}\label{eq:lsd}
        \frac{1}{n^q} \sum_{i=1}^{n^q} \delta_{\lambda_i(\mM)} (\dx) \aslim \frac{1}{q!} \, \fsc(x) \,  \dx + \lrp{1 - \frac{1}{q!}} \, {\delta_0}(\dx).
    \end{equation}
\end{corollary}
The proof follows the scheme exposed in Section~\ref{subsec:RMT}. The symmetric unfolding removes the deterministic null space created by repeated rows and columns. After the normalization by $\alpha_{p, q} \, n^{(p-2) / 4}$, the entries have the same leading variance as those of a standard Wigner matrix of dimension $n_q$; the remaining correlations are indexed by the finite equivalence classes~\eqref{eq:equivalence}. The proof shows, using Stein's lemma, that all dependencies other than the usual matrix transpose symmetry are asymptotically negligible in the resolvent equation.
\begin{remark}[Normalization Factor]
    One could expect a normalization by $\sqrt{n^q}$ given the resolvent matrix size ($n^q \times n^q$). The $n^{(p-2) / 4} = \sqrt{n^q} / \sqrt{n}$ factor we get instead comes from the $1 / \sqrt{n}$ factor in the definition of the noise model~\eqref{eq:noise_model}. Although slightly more cumbersome, we decide to keep this normalization in order to stick with the original work of~\citet{montanari_statistical_2014}. Several follow-up works indeed stick with this normalization, making the conjectured expression ``$n^{(p-2) / 4}$'' central in the Tensor PCA community~\citep{arous_landscape_2019,perry_statistical_2020,goulart_random_2022,arous_long_2023,lebeau_random_2025}, while some other remove the noise tensor normalization by $\sqrt{n}$ for conciseness~\citep{hopkins_tensor_2015,kim_community_2017,feldman_sharp_2023}, getting ``$n^{p/4}$''-like thresholds.
\end{remark}
\begin{remark}[Spectral Norm]\label{rem:spectral_norm}
    The weak convergence of the empirical spectral distribution towards the limiting spectral distribution is not sufficient to guarantee that no empirical eigenvalue ``escapes'' from the semicircle bulk support, which is crucial for estimation (see~\citealt{couillet_random_2022}, Section~2.3). In Appendix~\ref{subsec:spectral_norm}, we show that the spectral edge of the (normalized) symmetrically unfolded GOTE is the standard Wigner spectral edge~\citep{bai_necessary_1988}, by a high-moment argument adapted to the high-order dependencies~\eqref{eq:equivalence}.
\end{remark}

\subsection{Phase Transition}\label{subsec:transition}
The previous results characterize the deterministic equivalent and the limiting spectral distribution of the unfolded noise tensor $\tZ \sim \GOTE_{p, n}$. Coming back to the spiked tensor model $\tX = \lambda \, {\vv}^{\otimes p} + \tZ$~\eqref{eq:spiked_model}, we now study the recovery of the lifted signal $\vv^{\boxtimes q} \in \mbS^{n^q - 1}$ by eigendecomposition of the unfolded tensor $\mX \coloneqq \Mat_q(\tX)$. The next theorem establishes a sharp phase transition with precise overlaps, as a function of the SNR $\lambda$.
\begin{theorem}[Phase Transition]\label{thm:transition}
    Let $\tX = \lambda \, {\vv}^{\otimes p} + \tZ$, with $\lambda > 0$, $\vv \in \mbS^{n-1}$ and $\tZ \sim \GOTE_{p, n}$, $\mX \coloneqq \Phi_q(\tX) \in \Sym(n_q)$ and $(\hat\lambda, \hat \vv) \in \mbR_+\times\mbS^{n_q-1}$ the leading eigenpair of $\mX$. Writing $\tau \coloneqq \lambda / (\alpha_{p, q} \, n^{(p-2) / 4})$ and $\hat\tau \coloneqq \hat\lambda / (\alpha_{p, q} \, n^{(p-2) / 4})$, with $\alpha_{p, q} \coloneqq \sqrt{q! / (p - 1)!}$, one has
\begin{equation}\label{eq:phase_transition}
    \hat\tau \aslim
    \begin{cases}
        2 & \text{if } \tau \leq 1,\\
        \tau + \tau^{-1} & \text{if }  \tau > 1,
    \end{cases}
    \qquad \text{and} \qquad
    |{\hat \vv}\T \vv^{\odot q}|^2 \aslim
    \begin{cases}
        0  & \text{if } \tau \leq 1,\\
        1 - \tau^{-2}  & \text{if } \tau > 1,
    \end{cases}
\end{equation}
with $\vv^{\odot q} \coloneqq \mU\T \vv^{\boxtimes q} \in \mbS^{n_q - 1}$. These results apply verbatim to $\mX \coloneqq \Mat_q(\tX) \in \Sym(n^q)$.
\end{theorem}
The previous results generalize the GOE to higher orders. Indeed, when $p=2$, one has $q! = 1$ and $\alpha_{p, q} = 1$, therefore the empirical spectral distribution is precisely the Wigner semicircle distribution and the associated overlaps follow~\citep{benaych-georges_eigenvalues_2011}. We note that the support of the semicircle shrinks---as $\sqrt{q! / (p - 1)!}$---and the proportion of zero eigenvalues grows---as $1 - q!^{-1}$---with the tensor order $p$.
\begin{remark}[Standard Gaussian Noise]\label{rem:standard}
The previous results can be extended under the symmetric standard Gaussian noise model mentioned in Footnote~\ref{foot:standard}. We establish in Appendix~\ref{sec:standard} a similar phase transition, with $\lambda = \sqrt{q!} \, n^{p/4}$ instead of $\lambda = \sqrt{q! / (p - 1)!} \, n^{(p-2)/4}$.
\end{remark}
\begin{remark}[Multiple Spikes]
The previous results extend to finite-rank signals. In Appendix~\ref{sec:multispike}, we establish sharp phase transitions for partial recovery of the lifted planted signals and the subspace they span, under the \emph{canonical polyadic} (CP) model~\citep{hitchcock_expression_1927} with orthogonal spikes. We also discuss \emph{Tucker} models~\citep{tucker_mathematical_1966}.
\end{remark}
\begin{remark}[Exact Recovery]\label{rem:exact}
We see that the overlap tends quadratically to $1$ at the phase transition scale $\lambda \asymp n^{(p-2)/4}$. This means that such a simple algorithm as unfolding, which is commonly believed to ``break'' the tensor structure, can recover the planted signal $\vv$ for $\lambda = \omega(n^{(p-2) / 4})$, via a consistent rank-one extraction from the estimated lifted tensor $\vv^{\boxtimes q} \in \mbS^{n^q - 1}$. \emph{Recursive unfolding}, which applies an $n \times n^{q-1}$ unfolding to $\vv^{\boxtimes q}$, is one of those; it is conjectured by~\citet{montanari_statistical_2014} to undergo a sharp phase transition at $\lambda \asymp n^{(p-2) / 4}$, with \emph{exact} recovery. However, the techniques used in~\citet{montanari_statistical_2014} and \citet{feldman_sharp_2023} to prove this result in the asymmetric setting do not translate directly to the symmetric setting. We discuss this perspective in Appendix~\ref{sec:recursive} and conjecture a sharp phase transition from null to exact recovery at $\lambda = \alpha_{p, q} n^{(p-2) / 4}$.
\end{remark}

\subsection{Simulations}\label{subsec:simu}
\begin{figure}
    \centering
    \includegraphics[width=.7\linewidth]{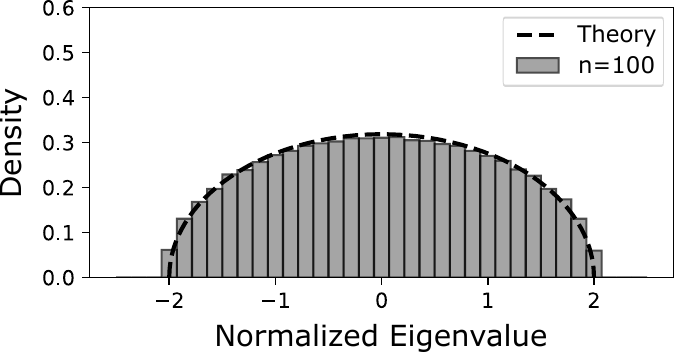}
    \caption{Theoretical and simulated spectral distribution of $\mM \coloneqq \alpha_{p, q}^{-1} \, n^{-(p-2) / 4} \, \Phi_q(\tZ)$, with $\tZ \sim \GOTE_{p, n}$, for $p=4$ and $n=100$ (see Theorem~\ref{thm:DE_symmetric}).}
    \label{fig:LSD}
\end{figure}

\begin{figure}
    \centering
    \includegraphics[width=.7\linewidth]{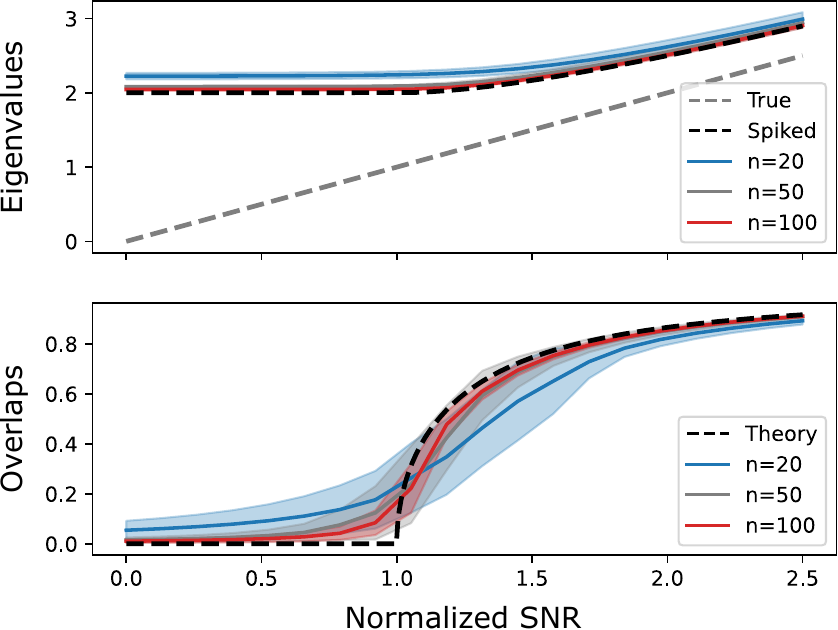}
    \caption{Theoretical and simulated eigenvalues and overlaps of $\mM \coloneqq \Phi_q(\lambda \, {\vv}^{\otimes p} + \tZ)$, with $\tZ \sim \GOTE_{p, n}$, for $p=4$ and $n \in \{20, 50, 100\}$ (see Theorem~\ref{thm:transition}).}
    \label{fig:overlap}
\end{figure}

To confront the asymptotic theoretical results with finite-sample observations, we consider a spiked tensor model $\tX = \lambda \, {\vv}^{\otimes p} + \tZ$~\eqref{eq:spiked_model} with $p=4$. We report the theoretical and simulated spectral distribution of the symetrically unfolded GOTE, $\mM \coloneqq \alpha_{p, q}^{-1} \, n^{-(p-2) / 4} \, \Phi_q(\tZ)$, in Figure~\ref{fig:LSD}, with $n=100$. Moreover, we report the eigenvalues and overlaps in Figure~\ref{fig:overlap}, with $n \in \{20, 50, 100\}$. For that experiment, we fix the signal $\vv \in \mbS^{n-1}$ via uniform sampling on the sphere, then, ten times independently, we sample $\tZ$ according to~\eqref{eq:noise_model} and increase $\tau \coloneqq \lambda / (\alpha_{p, q} \, n^{(p-2) / 4})$ linearly from $0$ to $2.5$.
We see that the empirical eigenvalues are well estimated by the limiting spectral distribution. Both leading eigenvalue and overlap are increasingly well estimated by the theoretical ones beyond phase transition, as $\lambda$ grows. The phase transition is notably already well captured for $n=50$.

\section{Discussion}\label{sec:discussion}
We have studied the spectral behavior of the unfolding algorithm under the spiked symmetric tensor model~\eqref{eq:spiked_model}. The strategy was to project the usual unfolding onto symmetric tensor subspaces, after which the matrix spectrally behaves like a Wigner matrix; this was rigorously proven using Stein's lemma and Cauchy--Schwarz's inequality on tensor slices with shared indices. We showed that, after normalization, the symmetric unfolding has a semicircular limiting spectrum and that its low-rank perturbation undergoes the usual BBP transition, with a sharp threshold $\lambda = \alpha_{p, q} \, n^{(p-2)/4}$ and an explicit overlap, settling the conjectured scaling of~\citet{montanari_statistical_2014}. Hence, despite its simplicity, the unfolding algorithm ranks among the best known polynomial-time estimators for symmetric Tensor PCA problems. To the best of our knowledge, this is the first random matrix analysis of the unfolding algorithm under the genuinely symmetric model of~\citet{montanari_statistical_2014} at that level of precision.

Several questions remain open. First, our analysis suggests a broader universality principle for tensor unfoldings. For a general order $p$ and matricization $\Mat_k(\mT)$, one can similarly project the lines and columns onto $\operatorname{Sym}^k(\mathbb R^n)$ and $\operatorname{Sym}^{p-k}(\mathbb R^n)$, obtaining a rectangular matrix $\Phi_k(\mT) \in \mbR^{n_k \times n_{p-k}}$. For $k<p-k$, a direct computation of the entrywise variance shows that, after normalization, the resulting model can be written in the form $n_{p-k}^{-1/2} \, \mX = \theta \, \vv^{\odot k} {\vv^{\odot p-k}}\T + n_{p-k}^{-1/2} \, \mZ$, with $\theta \coloneqq \lambda\sqrt{{(p-1)!} / ({k!\,n^{p-k-1}})}$, where $\mZ$ has ``mostly'' independent unit-variance entries. This is precisely the scaling of the disproportionate spiked matrix model of~\citet{feldman_spiked_2023}, with matrix dimensions $n_k \times n_{p-k}$. If the remaining dependencies induced by tensor symmetry~\eqref{eq:equivalence} are spectrally negligible, like in Lemma~\ref{lem:permut}, their Theorem~1 would therefore suggest a sharp transition at
\begin{equation*}
	\lambda_k = \gamma_{p, k} \, n^{(p-2)/4}, \qquad \text{with} \qquad \gamma_{p, k} \coloneqq \lrp{\frac{k! \, (p-k)!}{(p-1)!^2}}^{1/4}
\end{equation*}
and analogous formulas for the outlier singular value and the overlap with the left singular vector, corresponding to the shorter side. Interestingly, the polynomial order $n^{(p-2)/4}$ is independent of the unfolding, as in the asymmetric setting~\citep{arous_long_2023,feldman_sharp_2023}, but the sharp constant is \emph{not}: it is minimized by an ``as square as possible'' unfolding.\footnote{Indeed, we can easily show that the function $f\colon \{1, \ldots, p-1\} \ni q \mapsto q! (p-q)!$ decreases on $\{1, \ldots, \lfloor p/2 \rfloor\}$, (by studying $f(q+1)/f(q)$) and then increases on $\{\lceil p/2 \rceil, \ldots, p-1\}$ (by symmetry, $f(q) = f(p-q)$). Therefore it attains its minimum at $\lfloor p/2 \rfloor$ and $\lceil p/2 \rceil$ (so it has two global minima if and only if $p$ is odd).} Moreover, since the random matrix results hold under substantially more general assumptions than Gaussianity---see~\citet{benaych-georges_eigenvalues_2011} for square unfolding and~\citet{feldman_spiked_2023} for non-square unfolding---it is natural to expect that the same phenomenon extends to non-Gaussian symmetric noise under suitable moment and tail conditions. We conjecture that the symmetrically projected unfoldings indeed satisfy the corresponding disproportionate random matrix asymptotics; numerical simulations and additional comments are provided in Appendix~\ref{sec:choice_q}.

It would also be interesting to extend the present framework beyond the additive spiked model. Natural examples include \emph{moment} and \emph{cumulant} tensors~\citep{kunisky_tensor_2024,tang_detection_2026} arising in latent variable models~\citep{anandkumar_tensor_2014}. In each case, the main question is whether an appropriate symmetric unfolding reduces the problem to a tractable random matrix model with a sharp spectral transition.

Symmetric unfolding is related to other spectral approaches to Tensor PCA, notably the \emph{Kikuchi hierarchy}~\citep{wein_kikuchi_2025}. While symmetric unfolding exploits a single structured matricization, Kikuchi matrices introduce richer families of tensor transformations and provide a hierarchy of statistical and computational tradeoffs. Understanding the relation between these constructions, possibly through the recent random matrix and free-probabilistic techniques of~\citet{bandeira_matrix_2023,bandeira_matrix_2026}, could help understanding the impact of high-order symmetries on spectral transitions. Related questions arise for trace-invariant methods~\citep{ouerfelli_random_2022}, which exploit tensor-network contractions.

Finally, our results raise the more fundamental question of whether symmetry can significantly modify the spectral behavior. In the present setting, the dependencies created by high-order symmetries~\eqref{eq:equivalence} disappear at leading order after projection, resulting in a similar spectral behavior as the corresponding asymmetric model. This may not hold in general: localized signal priors or more-structured noise distributions may lead high-order dependencies to survive asymptotically, significantly impacting the eigenstructure.

\acks{The authors would like to thank Afonso S. Bandeira for insightful discussions around symmetric unfolding, Kikuchi matrices and intrinsic freeness. This work was supported by the French National Research Agency (ANR) via project LATENT (grant ANR-25-CE48-7508-01) and by the AI Interdisciplinary Institute ANITI funded by the France 2030 program (grant ANR-23-IACL-0002).}


\newpage

\appendix

\section{Proofs}\label{sec:proofs}

\subsection{Symmetric Unfolding (Proposition~\ref{prop:eigenstructure})}

\begin{proof}
By definition of $\mU$, one has $\Pi_{ij} \coloneqq [\mU \mU\T]_{ij} = \ind{\vj \in \mfS_{\vi}} / \#\mfS_{\vi}$. 
Moreover, as $\mM$ is hypersymmetric, every row and column is constant on permutation orbits, so $\mM = \mPi \mM \mPi = \mU \mU\T \mM \mU \mU\T$, with
\begin{equation*}
    [\mU\T \mM \mU]_{ij} = \sum_{\vi' \in \mfS_{\vi}} \sum_{\vj' \in \mfS_{\vj}} \frac{{M}_{i' j'}} {\sqrt{\#\mfS_{\vi} \#\mfS_{\vj}}} = \sum_{\vi' \in \mfS_{\vi}} \sum_{\vj' \in \mfS_{\vj}} \frac{{M}_{ij}} {\sqrt{\#\mfS_{\vi} \#\mfS_{\vj}}} = \sqrt{\#\mfS_{\vi} \#\mfS_{\vj}} \, {M}_{ij}.
\end{equation*}
Finally, since $\mU\T\mU=\mI_{n_q}$, the nonzero eigenpairs are transported by the isometric embedding $\mU$. The remaining $n^q-n_q$ dimensions belong to $\ker(\mM)$.
\end{proof}

\subsection{Deterministic Equivalent (Theorem~\ref{thm:DE_symmetric})}
\begin{proof}
The starting point is the resolvent identity $\mM \mQ = \mI_{n_q} + z \, \mQ$. Applying Stein's equality (Lemma~\ref{lem:stein}), one gets
    \begin{equation*}
        \expect{\mM \mQ}_{ij} = \sum_{k=1}^{n_q} \expect{{M}_{ik} {Q}_{kj}} = \sum_{k=1}^{n_q} \frac{p \, \#\mfS_{\vi} \, \#\mfS_{\vk}}{\alpha_{p, q}^2 \, n^{p/2} \, \#\mfS_{\vi\sqcup\vk}} \expect{\frac{\partial {Q}_{kj}}{\partial {M}_{ik}}},
    \end{equation*}
    with 
    \begin{equation}\label{eq:deriv}
        \frac{\partial {Q}_{kj}}{\partial {M}_{ik}} = - \lrb{\mQ \frac{\partial \mM}{\partial \mM_{ik}} \mQ}_{kj} = - \sum_{(\vi', \vk') \in \mathcal E(\vi,\vk)} {Q}_{\vk \vk'} {Q}_{\vi' \vj}.
    \end{equation}
    Noting that $\#\mfS_{\vi\sqcup\vk} = p! \#\mfS_{\vi} \#\mfS_{\vk} / q!^2$, for any ``generic'' $\vk \in \mathcal I_q$ such that $\vi\sqcap\vk = \varnothing$, one gets ${p \#\mfS_{\vi} \#\mfS_{\vk}} / {(\alpha_{p, q}^2 n^{p/2}\#\mfS_{\vi\sqcup\vk})} = q! / n^{p/2} \sim n_q^{-1}$. Consequently,
    \begin{equation}\label{eq:before_lemma}
        {\expect{\mM \mQ}}_{ij} \sim - \frac{1}{n_q} \sum_{\vk \in {\mathcal I_q}} \sum_{(\vi', \vk') \in \mathcal E(\vi,\vk)} \expect{{Q}_{\vk \vk'} {Q}_{\vi' \vj}}.
    \end{equation}
    Equation~\eqref{eq:before_lemma} is rather simple, but the sum over permutations of multi-indices it involves prevents us from obtaining a self-consistent equation on $\mQ$. Fortunately, we can show that the only permutation yielding non-negligible terms is the trivial one, matching $(\vi, \vk)$ with $(\vi', \vk')$; indeed, otherwise ${Q}_{\vk \vk'}$ and ${Q}_{\vi' \vj}$ have shared indices, leading the associated summation over $\vk$ to be negligible with respect to the right-hand side.
    \begin{lemma}[Negligible Permutations]\label{lem:permut}
        Let $s \in \{0, \ldots, q\}$ and $(\vi', \vk') \in \mathcal E(\vi,\vk)$ be such that $\vi' = (k_1, \ldots, k_s, i_{s+1}, \ldots, i_q)$ and $\vk' = (i_1, \ldots, i_s, k_{s+1}, \ldots, k_q)$. One has
        \begin{equation}
            \sum_{\vk \in {\mathcal I_q}} {Q}_{\vk \vk'} {Q}_{\vi' \vj} = O\lrp{n^{q-s} n^{\floor{s/2}}}.
        \end{equation}
    Here, $s$ represents the number of shared indices in $\vk \sqcup \vk'$ and $\vi' \sqcup \vj$.
    \end{lemma}
    \begin{proof}
        One has
        \begin{equation*}
            \sum_{\vk \in {\mathcal I_q}} {Q}_{\vk \vk'} {Q}_{\vi' \vj} = \sum_{(k_{s+1}, \ldots, k_q) \in \mathcal I_{q-s}} \langle{Q}_{\textcolor{cw6}{k_{s+1} \cdots k_q} \textcolor{cw3}{i_1 \cdots i_s} \textcolor{cw1}{k_{s+1} \cdots k_q}}, {Q}_{\textcolor{cw6}{i_{s+1} \cdots i_q} \textcolor{cw3}{j_1 \cdots j_s} \textcolor{cw1}{j_{s+1} \cdots j_q}}\rangle_\mathrm{F},
        \end{equation*}
        with the tensor slice notation
        \begin{align*}
                {Q}_{\textcolor{cw6}{k_{s+1} \cdots k_q} \textcolor{cw3}{i_1 \cdots i_s} \textcolor{cw1}{k_{s+1} \cdots k_q}} &\coloneqq \lrp{{Q}_{\textcolor{cw8}{k_1 \cdots k_s} \textcolor{cw6}{k_{s+1} \cdots k_q} \textcolor{cw3}{i_1 \cdots i_s} \textcolor{cw1}{k_{s+1} \cdots k_q}}}_{(\textcolor{cw8}{k_1, \ldots, k_s}) \in \mathcal I_{s}} \coloneqq \tA,\\
                {Q}_{\textcolor{cw6}{i_{s+1} \cdots i_q} \textcolor{cw3}{j_1 \cdots j_s} \textcolor{cw1}{j_{s+1} \cdots j_q}} &\coloneqq \lrp{{Q}_{\textcolor{cw8}{k_1 \cdots k_s} \textcolor{cw6}{i_{s+1} \cdots i_q} \textcolor{cw3}{j_1 \cdots j_s} \textcolor{cw1}{j_{s+1} \cdots j_q}}}_{(\textcolor{cw8}{k_1, \ldots, k_s}) \in \mathcal I_{s}} \coloneqq \tB.
        \end{align*}
        Let $\tA' \in \lrp{\mbR^n}^{\otimes s}$ be such that $A'_{a_1 \cdots a_s} = {A}_{a_1 \cdots a_s}$ if $(a_1, \ldots, a_s) \in \mathcal I_s$ and zero otherwise. Writing $\mA \coloneqq \Mat_{n^{\floor{s/2}}}(\tA')$, one has $\|\tA\|_\mathrm{F} = \|\mA\|_\mathrm{F} \leq \sqrt{n^{\floor{s/2}}} \|\mA\|_{\mathrm{op}} = O(\sqrt{n^{\floor{s/2}}})$. Similarly, one gets $\|\tB\|_\mathrm{F} = O(\sqrt{n^{\floor{s/2}}})$. 
        We conclude via the Cauchy--Schwarz inequality:
        \begin{equation*}
            \sum_{\vk \in {\mathcal I_q}} {Q}_{\vk \vk'} {Q}_{\vi' \vj} \leq \sum_{(k_{s+1}, \ldots, k_q) \in \mathcal I_{q-s}} \|\tA\|_\mathrm{F} \|\tB\|_\mathrm{F} = O\lrp{n^{q-s} n^{\floor{s/2}}}.
        \end{equation*}
    \end{proof}
    Hence, as soon as $\vk$ and $\vi'$ share one index, the associated sum over $\vk$ is $O(n^{q-s} n^{\floor{s/2}}) = O(n^q n^{-\ceil{s/2}}) = O(n^{q-1})$. Consequently, one has 
    \begin{equation}
        \sum_{\vk \in {\mathcal I_q}} \sum_{(\vi', \vk') \in \mathcal E(\vi,\vk)} {{Q}_{\vk \vk'} {Q}_{\vi' \vj}} \sim \sum_{\vk \in {\mathcal I_q}} {{Q}_{\vk \vk} {Q}_{\vi \vj}}.
    \end{equation}    
    Lemma~\ref{lem:permut} implies that  high-order dependencies are negligible, from the expected resolvent's point of view. Applying it to the equation~\eqref{eq:before_lemma} yields 
    \begin{equation*}
        {\expect{\mM \mQ}}_{ij} \sim - {n_q}^{-1} \sum_{\vk \in {\mathcal I_q}} \expect{{Q}_{\vk \vk} {Q}_{\vi \vj}}.
    \end{equation*}
    Writing $s(z) \coloneqq {n_q}^{-1} \tr{\mQ}$ and using $\mM \mQ = \mI_{n_q} + z \, \mQ$, we get
    \begin{equation}\label{eq:sc_resolvent}
        \mI_{n_q} + z \, \expect{\mQ} + \expect{s(z) \, \tQ} = o(1).
    \end{equation}
    Taking the normalized trace yields, as $n \to \infty$, the self-consistent equation $s^2 + z\, s + 1 = 0$, from which we recognize the classical Stieltjes transform of the semicircular distribution.
    
    Substituting this relation back into~\eqref{eq:sc_resolvent} yields, for every deterministic $\mA, \va, \vb$ such that $\|\mA\|_{\mathrm{op}} = \|\va\|_2 = \|\vb\|_2 =  1$,
    $\frac1{n_q}\tr{\mA\lrp{\expect{\mQ}-\ssc(z)\mI}} \to 0$ and $\va\T\lrp{\expect{\mQ}-\ssc(z)\mI}\vb \to 0$.

    Hence, $\bar \mQ \coloneqq \ssc \, \mI_{n_q}$ is a good candidate for deterministic equivalence of $\mQ$ (Definition~\ref{def:DE}). However, the previous results hold in expectation; to show concentration, we apply the Nash--Poincaré inequality (Lemma~\ref{lem:np}) to the functions $f(\vx) = \tr{\mA \mQ}$ and $g(\vx) =\va\T \mQ \vb$, where $\vx \coloneqq \operatorname{vec}\bigl({M}_{\vi}\bigr)_{\vi \in \mathcal I_p}$. 
    Using again the derivative~\eqref{eq:deriv}, we get
    \begin{align*}
        \mbV\lrb{\tr{\mA \mQ}} &= O\lrp{\frac{1}{n^q} \sum_{\vi \in \mathcal I_p} \expect{\lrp{\frac{\partial\tr{\mA\mQ}}{\partial{M}_{\vi}}}^2}}\\
        &= O\lrp{\frac{1}{n^q} \sum_{\vi \in \mathcal I_p} \expect{\lrp{\sum_{k, l \in [n^q]} {A}_{kl} \frac{ \partial{Q}_{lk}}{\partial{M}_{\vi}}}^2}}\\
        &= O\lrp{\frac{1}{n^q} \sum_{\vi \in \mathcal I_p} \expect{\lrp{\sum_{\vi'_1 \sqcup \vi'_2 = \vi}\sum_{k, l \in [n^q]} {A}_{kl} {Q}_{i'_1 k} {Q}_{l i'_2}}^2}}\\
        &= O\lrp{\frac{1}{n^q} \sum_{\vi \in \mathcal I_p} \expect{\sum_{\vi'_1 \sqcup \vi'_2 = \vi} \lrp{\sum_{k, l \in [n^q]} {A}_{kl} {Q}_{i'_1 k} {Q}_{l i'_2}}^2}}\\
        &= O\lrp{\frac{1}{n^q} \sum_{\vi_1, \vi_2 \in \mathcal I_q} \expect{\sum_{\vi'_1 \sqcup \vi'_2 = \vi_1 \sqcup \vi_2} \lrp{\sum_{k, l \in [n^q]} {A}_{kl} {Q}_{i'_1 k} {Q}_{l i'_2}}^2}}\\
        &= O\lrp{\frac{1}{n^q} \binom{2q}{q} \expect{\sum_{\vi_1, \vi_2 \in \mathcal I_q} \lrp{\sum_{k, l \in [n^q]} {Q}_{i_1 k} {A}_{kl} {Q}_{l i_2}}^2}}\\
        &= O\lrp{\expect{\|\mQ \mA \mQ\|_\mathrm{F}^2} / n^q}\\
        &= O\lrp{\expect{n^q \|\mQ\|_\mathrm{op}^2 \|\mA\|_\mathrm{op}^2 \|\mQ\|_\mathrm{op}^2} / n^q}\\
        &= O(1).
    \end{align*}
    Noticing that $\va\T \mQ \vb = \tr{\vb \va\T\mQ}$, we can apply the same sequence of equalities for the concentration of bilinear forms of the resolvent. Writing ${A}_{kl} = b_k a_l$, we get
    \begin{align}
        \mbV\lrb{\va\T \mQ \vb} &= O\lrp{\frac{1}{n^q} \sum_{\vi \in \mathcal I_p} \expect{\sum_{\vi'_1 \sqcup \vi'_2 = \vi} \lrp{\sum_{k, l \in [n^q]} b_k a_l {Q}_{i'_1 k} {Q}_{l i'_2}}^2}}\\
        &= O\lrp{\expect{\|\mQ \va\|_2^2 \|\mQ \vb\|_2^2} / n^q}\\
        &= O\lrp{\expect{\|\mQ\|_\mathrm{op}^2 \|\mQ\|_\mathrm{op}^2} / n^q}\\
        &= O(1/n^q).
    \end{align}
    Hence, one has $\mbV\lrb{\frac{1}{n^q} \tr{\mA \lrp{\mQ - \bar \mQ}}} = O(n^{-p})$ and $\mbV\lrb{\va\T \lrp{\mQ - \bar\mQ}\vb} = O(n^{-q})$. The variance bounds are both summable for $q \geq 2$, therefore the convergence is almost sure by virtue of Chebyshev's inequality and Borel--Cantelli's lemma. For $q=1$, the matrix is an ordinary GOE and the almost-sure resolvent convergence is classical (see, for instance, Theorem~2.5 of~\citealt{couillet_random_2022}). 
    
    This proves the deterministic equivalent~\eqref{eq:resolvent_symmetric}. The almost-sure convergence of the Stieltjes transform to $\ssc$ follows. The Stieltjes continuity theorem then gives weak convergence to the standard semicircle law~\eqref{eq:lsd_symmetric}.
\end{proof}

\begin{proof}[Proof of Corollary~\ref{cor:DE}]
The full unfolding satisfies $\alpha_{p, q} \, n^{(p-2)/4} \, \mM = \mU \Phi_q(\tZ) \mU\T$ on $\operatorname{Im}(\mU)$ and is zero on its orthogonal complement. Since $n_q / n^q \to 1/q!$, its empirical law is the mixture of the semicircle scaled by $\alpha_{p, q}$ and a Dirac mass of weight $1 - 1/q!$.
\end{proof}

\subsection{Spectral Norm (Remark~\ref{rem:spectral_norm})}\label{subsec:spectral_norm}
We establish the spectral norm of the symmetrically unfolded GOTE by adapting the standard high-moment argument (see Section~2.3 of~\citealt{tao_topics_2012}) and showing that the additional identifications induced by tensor symmetry are negligible.
\begin{proposition}[Spectral Norm]\label{prop:spectral_norm}
    Let $\tZ \sim \GOTE_{p, n}$ and $\mM \coloneqq \alpha_{p,q}^{-1} \, n^{-(p-2) / 4} \, \Phi_q(\tZ)$, with $\alpha_{p,q} \coloneqq \sqrt{q! / (p - 1)!}$. One has $\|\mM\|_{\mathrm{op}} \aslim 2$.   
\end{proposition}
\begin{proof}
Expanding the trace and applying Wick's formula gives
\begin{equation*}
\expect{\operatorname{tr}(\mM^{2k})} = \sum_{\pi\in\mathcal P_2(2k)} \sum_{\vi_1,\ldots,\vi_{2k}\in\mathcal I_q} \prod_{\{a,b\}\in\pi} \expect{\mM_{\vi_a\vi_{a+1}} \mM_{\vi_b\vi_{b+1}}},
\end{equation*}
where $\mathcal P_2(2k)$ denotes the set of pair partitions of $[2k]$ and $\vi_{2k+1}=\vi_1$. A covariance is nonzero only if $\vi_a\sqcup\vi_{a+1} = \vi_b\sqcup\vi_{b+1}$. The \emph{standard} identifications are $(\vi_a, \vi_{a+1}) = (\vi_b, \vi_{b+1})$ and $(\vi_a, \vi_{a+1}) = (\vi_{b+1}, \vi_b)$; all other identifications are called \emph{non-standard}. 

We first show that, for every $\varepsilon>0$, uniformly for $k\le c_\varepsilon\ln n$,
\begin{equation}\label{eq:moment_eps}
\mathbb E[\operatorname{tr}(\mM^{2k})] \le n_q(2+\varepsilon/2)^{2k}.
\end{equation}
To this end, write $W(\pi,\vi_1,\ldots,\vi_{2k}) \coloneqq \prod_{\{a,b\}\in\pi} \mathbb E[\mM_{\vi_a\vi_{a+1}}\mM_{\vi_b\vi_{b+1}}]$. The covariance structure of the symmetric unfolding gives, for some constants $A_q, B_q$,
\begin{equation}\label{eq:moment_graph_count}
\sum_{\substack{\pi,\vi_1,\ldots,\vi_{2k}\\\text{non-standard}}} W(\pi,\vi_1,\ldots,\vi_{2k}) \le \frac{A_q k^{B_q}}{n} \sum_{\substack{\pi,\vi_1,\ldots,\vi_{2k}\\ \text{standard}}} W(\pi,\vi_1,\ldots,\vi_{2k}).
\end{equation}
Indeed, each non-standard identification leads to a factor $O(n^{-1})$ with respect to the associated standard identification---see Lemma~\ref{lem:permut}---and the $k^{B_q}$ factor accounts for the possible choices of locations of the non-standard covariances. The standard Wigner moment count gives, uniformly for $k\le c_\varepsilon\ln n$,
\begin{equation*}
    \sum_{\substack{\pi,\vi_1,\ldots,\vi_{2k}\\\text{standard}}} W(\pi,\vi_1,\ldots,\vi_{2k}) \le n_q(2+\varepsilon/4)^{2k}.
\end{equation*}
Together with~\eqref{eq:moment_graph_count}, this proves~\eqref{eq:moment_eps} for all sufficiently large $n$.

Now let $k_n=\lfloor c_\varepsilon\ln n\rfloor$, with $c_\varepsilon$ sufficiently large. Since $\|\mM\|_{\mathrm{op}}^{2k_n} \le\operatorname{tr}(\mM^{2k_n})$, Markov's inequality and~\eqref{eq:moment_eps} give
\begin{equation*}
    \mathbb P\!\left(\|\mM\|_{\mathrm{op}}>2+\varepsilon\right) \le n_q
    \left(\frac{2+\varepsilon/2}{2+\varepsilon}\right)^{2k_n}.
\end{equation*}
Since $n_q = O(n^q)$, $c_\varepsilon$ can be chosen so that the right-hand side is summable in $n$. Hence, by Borel--Cantelli, $\limsup_{n\to\infty}\|\mM\|_{\mathrm{op}}\le2$, almost surely. Finally, the empirical spectral distribution converges almost surely to the semicircle law. Since the latter has positive mass in $(2-\varepsilon,2)$, almost surely $\mM$ has an eigenvalue larger than $2-\varepsilon$ for all sufficiently large $n$. Thus $\liminf_{n\to\infty}\|\mM\|_{\mathrm{op}}\ge2-\varepsilon$. Letting $\varepsilon\downarrow0$ concludes that $\|\mM\|_{\mathrm{op}} \to 2$, almost surely.
\end{proof}
\subsection{Phase Transition (Theorem~\ref{thm:transition})}
\begin{proof}
Similarly to Theorem~\ref{thm:DE_symmetric}, let $\mM \coloneqq \alpha_{p, q}^{-1} \, n^{-(p-2) / 4} \, \Phi_q(\tZ)$ and $\mQ(z) \coloneqq \lrp{\mM - z \, \mI_{n^q}}^{-1}$. By Proposition~\ref{prop:spectral_norm}, $\lambda_{\max}(\mM) \to 2$ almost surely.

Suppose first that $\tau>1$ and let $z \in \mathbb C\setminus[-2,2]$. Then, 
\begin{equation}\label{eq:det}
    \det(\mX - z \, \mI) = \det(\mM-z\, \mI)\bigl(1+\tau\,{\vv^{\odot q}}^\top\mQ(z){\vv^{\odot q}}\bigr).
\end{equation}
By Theorem~\ref{thm:DE_symmetric}, one has ${\vv^{\odot q}}\T \mQ(z) \vv^{\odot q} \to \ssc(z)$. Since $1+\tau \ssc(z)=0$ has a unique solution outside $[-2,2]$, that is $\rho(\tau) \coloneqq \tau+\tau^{-1}$, $\mX$ has a unique eigenvalue $\hat\tau$ converging to $\rho(\tau) > 2$. Let $\Gamma$ be a small contour around $\rho(\tau)$. By Cauchy’s integral formula (Lemma~\ref{lem:resolvent}) and Sherman--Morrison's formula, one has 
\begin{equation*}
    |{\hat \vv}\T \vv^{\odot q}|^2 = -\frac{1}{2\pi \ii} \oint_{\Gamma} \frac{{\vv^{\odot q}}\T \mQ(z) {\vv^{\odot q}}}{1 + \tau \, {\vv^{\odot q}}\T \mQ(z) {\vv^{\odot q}}} \longrightarrow -\frac1{2\pi\ii} \oint_\Gamma \frac{\ssc(z)} {1+\tau \ssc(z)}\,\dz.
\end{equation*}
The integrand has a unique simple pole at $\rho(\tau)$, so the residue
theorem yields
\begin{equation*}
    |{\hat \vv}\T \vv^{\odot q}|^2 \longrightarrow -\frac{\ssc(\rho(\tau))} {\tau \ssc'(\rho(\tau))} = 1 - \frac{1}{\tau^2}.
\end{equation*}
Now suppose that $\tau\le1$ and fix $\varepsilon>0$. For $z\ge2+\varepsilon$, one has $1+\tau \ssc(z)\ge 1 + \ssc(2+\varepsilon)>0$. By Theorem~\ref{thm:DE_symmetric}, $1+\tau\,{\vv^{\odot q}}^\top\mQ(z){\vv^{\odot q}}$ is therefore bounded away from zero, hence the
determinant identity~\eqref{eq:det} shows that $\mX$ has no eigenvalue larger
than $2+\varepsilon$. By Weyl's inequality, $
\lambda_{\max}(\mX) \ge \lambda_{\max}(\mM) \to 2$; eventually, $\lambda_{\max}(\mX)\to 2$. Finally, the subcritical finite-rank deformation result for Wigner
matrices, together with the isotropic deterministic equivalent proved
in Theorem~\ref{thm:DE_symmetric}, gives $|{\hat \vv}\T \vv^{\odot q}|^2 \to 0$.

The results apply similarly to $\Mat_q(\tX)$, as ${(\mU \hat \vv)}\T (\mU \vv^{\odot q}) = {\hat \vv}\T \mU\T \mU \vv^{\boxtimes q} = {\hat \vv}\T \vv^{\odot q}$.
\end{proof}
   
\section{Standard Gaussian Noise}\label{sec:standard}
\begin{figure}
    \centering
    \includegraphics[width=.8\linewidth]{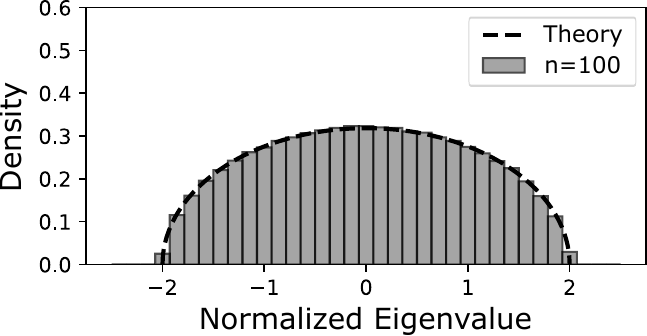}
    \caption{Theoretical and simulated spectral distribution of the normalized symmetrically unfolded standard Gaussian noise, $\mM \coloneqq \beta^{-1} \, n^{- p / 4} \, \Phi_q(\tZ)$, with $\tZ \sim \SG_{p, n}$, for $p=4$ and $n=100$ (see Theorem~\ref{thm:DE_symmetric_standard}).}
    \label{fig:LSD_standard}
\end{figure}

\begin{figure}
    \centering
    \includegraphics[width=.8\linewidth]{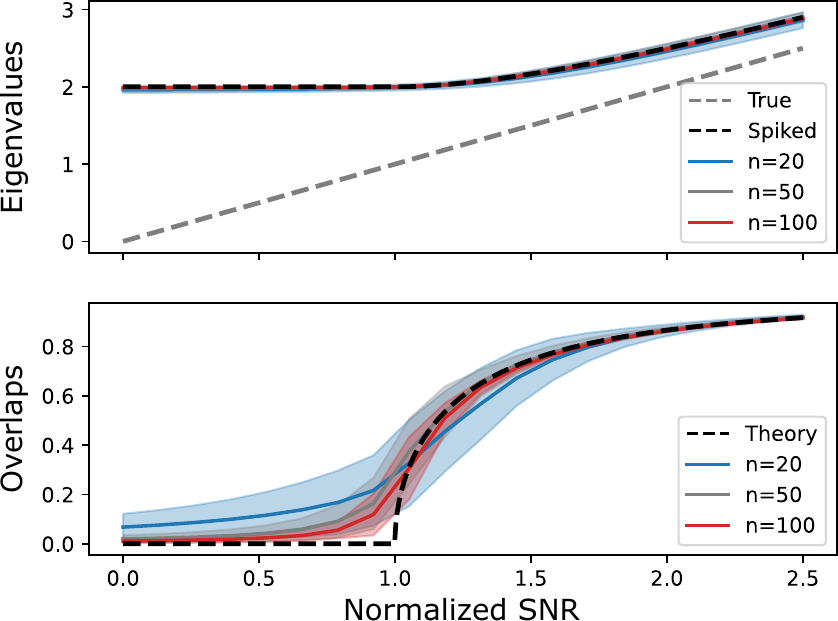}
    \caption{Theoretical and simulated eigenvalues and overlaps for the symmetrically unfolded spiked tensor $\mM \coloneqq \Phi_q(\lambda \, {\vv}^{\otimes p} + \tZ)$, with $\tZ \sim \SG_{p, n}$, for $p=4$ and $n \in \{20, 50, 100\}$ (see Theorem~\ref{thm:transition_standard}).}
    \label{fig:overlap_standard}
\end{figure}

This section extends the main text results to the spiked tensor model $\tX \coloneqq \lambda \, \vv^{\otimes p} + \tZ$, where $\tZ \sim \SG_{p, n}$ has independent standard Gaussian entries up to tensor symmetry:
\begin{equation}
    \begin{cases}
        {Z}_{i_1 \cdots i_p} &\simiid \mcN(0, 1),\\
        {Z}_{i_{\sigma(1)}  \cdots i_{\sigma(p)}} &= {Z}_{i_1 \cdots i_p}.
    \end{cases}
\end{equation}
\begin{theorem}[Deterministic Equivalent]\label{thm:DE_symmetric_standard}
    Let $\tZ \sim \SG_{p, n}$, $\mM \coloneqq \beta^{-1} \, n^{-p/4} \, \Phi_q(\tZ)$ and $\mQ(z) \coloneqq \lrp{\mM - z \, \mI_{n_q}}^{-1}$, with $\beta \coloneqq \sqrt{q!}$. Then,
    \begin{equation}\label{eq:resolvent_symmetric_standard}
        \mQ(z) \longleftrightarrow \ssc(z) \, \mI_{n_q}, \qquad \text{with} \qquad \ssc(z) \coloneqq - \bigl(z - \sqrt{z^2 - 4}\bigr) / 2,
    \end{equation}
   in the sense of Definition~\ref{def:DE}. Consequently, the empirical spectral distribution of $\mM$ converges weakly almost surely to the standard Wigner semicircle law:
    \begin{equation}\label{eq:lsd_symmetric_standard}
        \frac{1}{n_q}\sum_{i=1}^{n_q}\delta_{\lambda_i(\mM)} \aslim \fsc(x)\,\dx, \qquad \text{with} \qquad  \fsc(x) = \frac{\sqrt{(4-x^2)^+}}{2\pi}.
    \end{equation}
\end{theorem}
\begin{corollary}[Standard Unfolding]\label{cor:DE_standard}
    Let $\tZ \sim \SG_{p, n}$, $\mM \coloneqq \beta^{-1} \, n^{-p/4} \, \Mat_q(\tZ)$ and $\mQ(z) \coloneqq \lrp{\mM - z \, \mI_{n^q}}^{-1}$, with $\beta \coloneqq \sqrt{q!}$. Then,
    \begin{equation}\label{eq:resolvent_standard}
        \mQ(z) \longleftrightarrow \ssc(z) \, \mPi + s_{\delta_0}(z) \, (\mI_{n^q} - \mPi), \qquad \text{with} \qquad s_{\delta_0}(z) \coloneqq -\frac{1}{z},
    \end{equation}
    in the sense of Definition~\ref{def:DE}. Consequently, the empirical spectral distribution of $\mM$ converges weakly almost surely to a mixture of a semicircle and a Dirac:
    \begin{equation}\label{eq:lsd_standard}
        \frac{1}{n^q} \sum_{i=1}^{n^q} \delta_{\lambda_i(\mM)} (\dx) \aslim \frac{1}{q!} \, \fsc(x) \,  \dx + \lrp{1 - \frac{1}{q!}} \, {\delta_0}(\dx).
    \end{equation}
\end{corollary}
\begin{theorem}[Phase Transition]\label{thm:transition_standard}
    Let $\tX = \lambda \, {\vv}^{\otimes p} + \tZ$, with $\lambda > 0$, $\vv \in \mbS^{n-1}$ and $\tZ \sim \SG_{p, n}$, $\mX \coloneqq \Phi_q(\tX) \in \Sym(n_q)$ and $(\hat\lambda, \hat \vv) \in \mbR_+\times\mbS^{n_q-1}$ the leading eigenpair of $\mX$. Writing $\tau \coloneqq \lambda / (\beta \, n^{p/4})$ and $\hat\tau \coloneqq \hat\lambda / (\beta \, n^{p/4})$, with $\beta \coloneqq \sqrt{q!}$, one has
\begin{equation}\label{eq:phase_transition_standard}
    \hat\tau \aslim
    \begin{cases}
        2 & \text{if } \tau \leq 1,\\
        \tau + \tau^{-1} & \text{if }  \tau > 1,
    \end{cases}
    \qquad \text{and} \qquad
    |{\hat \vv}\T \vv^{\odot q}|^2 \aslim
    \begin{cases}
        0  & \text{if } \tau \leq 1,\\
        1 - \tau^{-2}  & \text{if } \tau > 1,
    \end{cases}
\end{equation}
with $\vv^{\odot q} \coloneqq \mU\T \vv^{\boxtimes q} \in \mbS^{n_q - 1}$. These results apply verbatim to $\mX \coloneqq \Mat_q(\tX) \in \Sym(n^q)$.
\end{theorem}
\begin{proof}[Theorem~\ref{thm:DE_symmetric_standard}, Corollary~\ref{cor:DE_standard} and Theorem~\ref{thm:transition_standard}]
The proof is identical in structure to the one of $\tZ \sim \GOTE_{p, n}$. After multiplication by the symmetric basis weights and division by $\beta \, n^{p/4}$, a generic entry of $\mM$ has variance $q!^2 / ({\beta^2 n^{p/2}}) = n_q^{-1} \, ({1+O(n^{-1})})$. The dependency classes are exactly the same, whence the negligibility of non-standard symmetries. Eventually, the Stein, Nash--Poincaré and Sherman--Morrison arguments apply similarly.
\end{proof}
We report the theoretical and simulated spectral distribution of the symmetrically unfolded standard Gaussian model, $\mM \coloneqq \beta^{-1} \, n^{- p / 4} \, \Phi_q(\tZ)$, in Figure~\ref{fig:LSD_standard}, with $n=100$.
Moreover, we report the eigenvalues and overlaps in Figure~\ref{fig:overlap_standard}, with $n \in \{20, 50, 100\}$, in a similar experimental setting to Section~\ref{subsec:simu}.
We see that the empirical eigenvalues are well estimated by the limiting spectral distribution. Both leading eigenvalue and overlap are increasingly well estimated by the theoretical ones beyond phase transition, as $\lambda$ grows.

The match between theory and simulations is slightly better under $\tZ \sim \SG_{p, n}$ than $\tZ \sim \GOTE_{p, n}$; see Figures~\ref{fig:LSD} and~\ref{fig:overlap} for comparison. This may be due to the extra asymptotic approximations that are performed in the proof under the GOTE noise, notably $\#\mfS_{\vi\sqcup\vk} = p! \#\mfS_{\vi} \#\mfS_{\vk} / q!^2$, which holds only for generic indices.

\section{Finite-Rank Extensions}\label{sec:multispike}
\begin{figure}
    \centering
    \includegraphics[width=\linewidth]{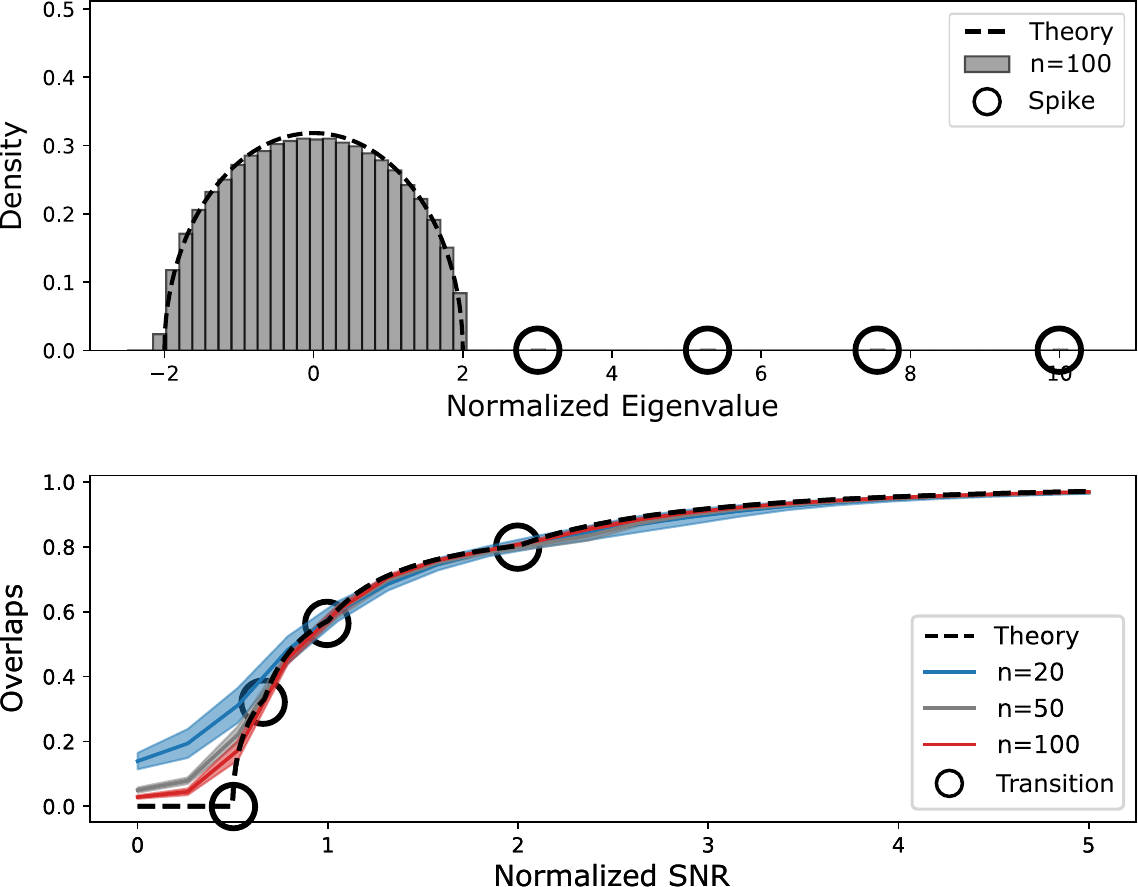}
    \caption{Theoretical and simulated spectral distribution and subspace overlap of the symmetrically unfolded CP model $\mM \coloneqq \alpha_{p, q}^{-1} \, n^{- (p - 2) / 4} \, \Phi_q(\sum_{k=1}^r\lambda_k\vv_k^{\otimes p}+\tZ)$, with $\tZ \sim \GOTE_{p, n}$, for $p=4$, $r=4$ and $n=100$ (see Corollary~\ref{cor:multispike}).}
    \label{fig:multispike}
\end{figure}

Theorem~\ref{thm:transition} on phase transition extends to a finite number of orthogonal planted signals.
\begin{corollary}[Multiple Spikes]
\label{cor:multispike}
    Consider the CP model
    \begin{equation}\label{eq:CP}
        \tX=\sum_{k=1}^r\lambda_k\vv_k^{\otimes p}+\tZ,
    \end{equation}
    where $\lambda_1>\cdots>\lambda_r>0$, $\vv_1,\ldots,\vv_r\in\mathbb S^{n-1}$ are orthonormal and $\tZ\sim\GOTE_{p, n}$. Let $\mX=\Phi_q(\tX)\in\Sym(n_q)$ and let $(\hat\lambda_k,\hat\vv_k)_{k=1}^r$ denote its $r$ leading eigenpairs, ordered decreasingly. Writing $\tau_k \coloneqq \lambda_k / (\alpha_{p, q} \, n^{(p-2) / 4})$ and $\hat\tau_k \coloneqq \hat\lambda_k / (\alpha_{p, q} \, n^{(p-2) / 4})$, with $\alpha_{p, q} \coloneqq \sqrt{q! / (p - 1)!}$, one has, for every fixed $k\in{1,\ldots,r}$,
    \begin{equation*}
        \hat\tau_k \aslim
        \begin{cases}
            2,&\tau_k\le1,\\
            \tau_k+\tau_k^{-1},&\tau_k>1,
        \end{cases}
        \qquad \text{and} \qquad
        \bigl|\hat\vv_k\T\vv_k^{\odot q}\bigr|^2 \aslim
        \begin{cases}
            0,&\tau_k\le1,\\
            1-\tau_k^{-2},&\tau_k>1.
        \end{cases}
    \end{equation*}
    Moreover, for $l\neq k$, $\bigl|\hat\vv_k\T\vv_l^{\odot q}\bigr| \to 0$. Consequently, writing 
    $\mV =
    \begin{bmatrix}
    \vv_1^{\odot q}&\cdots&\vv_r^{\odot q}
    \end{bmatrix}
    $
    and 
    $\hat\mV = 
    \begin{bmatrix}
    \hat\vv_1&\cdots&\hat\vv_r
    \end{bmatrix}$, one has
    \begin{equation}\label{eq:multispike_subspace_overlap}
        \|\hat\mV\T\mV\|_F^2 \aslim \sum_{k=1}^r(1-\tau_k^{-2})^+.
    \end{equation}
    These results apply verbatim to $\mX \coloneqq \Mat_q(\tX) \in \Sym(n^q)$.
\end{corollary}
\begin{proof}
After the symmetric unfolding and normalization, the model is $\mM +\sum_{k=1}^r \tau_k \vu_k \vu_k\T$, with $\vu_k=\vv_k^{\odot q}$. The vectors $\vu_k$ are orthonormal because $\langle\vu_k,\vu_l\rangle=\langle\vv_k,\vv_l\rangle^q$.
The deformation has fixed rank. The isotropic deterministic equivalent from Theorem~\ref{thm:DE_symmetric} reduces its outlier equations to $1 + \tau_k \, \ssc(z) = 0$ after applying \emph{Woodbury's formula}---a rank-$r$ version of Sherman--Morrison's. Distinct supercritical strengths therefore give separated outliers at $\tau_k+\tau_k^{-1}$ and the residue computation is the same as in the rank-one proof, yielding $|\hat\vu_k\T\vu_k|^2 \to (1-\tau_k^{-2})^+$ and vanishing cross-overlaps. The following result on subspace alignment follows from the convergence of the overlap matrix.
\end{proof}
We report the theoretical and simulated spectral distribution and subspace overlap of the symmetrically unfolded CP model~\eqref{eq:CP}, $\mM \coloneqq \alpha_{p, q}^{-1} \, n^{- (p - 2) / 4} \, \Phi_q(\tX)$, in Figure~\ref{fig:multispike}, with $p=4$, $r=4$ and $n\in\{20, 50, 100\}$.
We see that the $r$ largest eigenvalues escape the support of the limiting spectral density; they are therefore above the phase transition threshold. The theoretical subspace overlap curve undergoes multiple phase transitions, one for each spike. It increasingly well estimates the finite-sample simulations as $n$ and $\lambda$ grow.

The preceding subspace recovery result suggests an extension to other tensor models, such as the \emph{Tucker decomposition}~\citep{tucker_mathematical_1966}:
\begin{equation}\label{eq:symmetric_tucker}
    \tX = \sum_{k_1,\ldots,k_p=1}^r H_{k_1\cdots k_p} \, (\vv_{k_1}\otimes\cdots\otimes\vv_{k_p}) + \tZ,
\end{equation}
where $\tH\in\Sym^p(\mathbb R^r)$, $\vv_1, \ldots, \vv_r \in \mbS^{n-1}$ are orthonormal and $\tZ\sim\GOTE_{p, n}$. Such a model is notably relevant in applications where the signal is described by a low-dimensional subspace rather than rank-one components~\citep{sidiropoulos_tensor_2017,lebeau_random_2025}. However, the resulting \emph{lifted subspace} estimate---an $\binom{n+q-1}{q} \times \binom{r+q-1}{q}$ matrix---would not straightforwardly lead to an estimate of the subspace $\operatorname{span}(\vv_1, \ldots, \vv_r)$. We therefore leave the Tucker extension for future research.

\section{Recursive Unfolding}\label{sec:recursive}
\begin{figure}
    \centering
    \includegraphics[width=.8\linewidth]{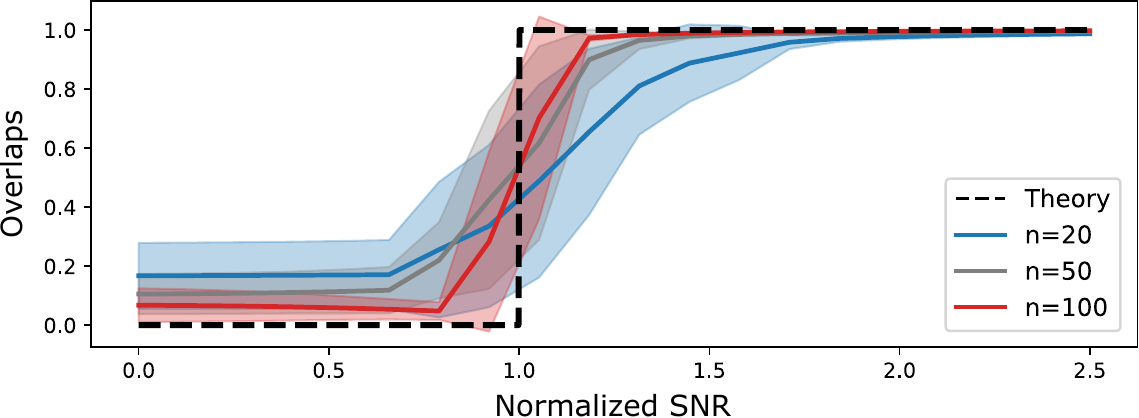}
    \caption{Theoretical and simulated overlaps for the recursively unfolded spiked tensor $\tT \coloneqq \lambda \, {\vv}^{\otimes p} + \tZ$, with $\tZ \sim \GOTE_{p, n}$, for $p=4$ and $n \in \{20, 50, 100\}$ (see Conjecture~\ref{conj:recursive}).}
    \label{fig:recursive}
\end{figure}

The main text analysis concerns a single unfolding of the tensor, resulting in overlap formulas with the \emph{lifted} vector $\vv^{\boxtimes q} \in \mbS^{n^q - 1}$. It is natural to ask whether the corresponding recursive unfolding procedure, introduced and analyzed in the asymmetric setting by~\citet[Theorem~5]{montanari_statistical_2014}, also achieves exact recovery in the genuinely symmetric model. The similarity between the two Gaussian tensor models---notably enabled by Lemma~\ref{lem:permut} on the negligible high-order symmetries---suggests that it should.
\begin{conjecture}[Recursive Unfolding]\label{conj:recursive}
Let $\tX=\lambda\,\vv^{\otimes p}+\tZ$, with $\vv\in\mbS^{n-1}$ and $\tZ\sim\GOTE_{p, n}$. Let $\mX = \Mat_q(\tX)\in\Sym(n^q)$ and denote by $\hat\vu\in\Sym^q(\mathbb R^n)$ a leading eigenvector of $\mX$. Let $\hat\vv\in\mbS^{n-1}$ be a leading left singular vector of the first-mode matricization $\Mat_1(\hat\vu)\in\mathbb R^{n\times n^{q-1}}$. Writing $\tau\coloneqq {\lambda} / {(\alpha_{p, q} n^{(p-2)/4})}$ with $\alpha_{p, q}\coloneqq\sqrt{{q!} / {(p-1)!}}$, one has
\begin{equation}\label{eq:recursive}
|\hat\vv^\top\vv|^2 \aslim
\begin{cases}
0 & \text{if } \tau\leq 1\\
1 & \text{if } \tau>1,
\end{cases}
\end{equation}
\end{conjecture}
\begin{proof}[Idea of proof]
The supercritical case can be understood by separating the signal component of the leading eigenvector from its orthogonal remainder. Write 
\begin{equation}
    \hat\vu = \rho_n \,\vu + \sqrt{1-\rho_n^2} \, \vg,
\end{equation}
with $\vu=\vv^{\boxtimes q}$, $\vg\perp\vu$ and $\|\vg\|=1$  and set $\mG=\Mat_1(\vg)$. By Theorem~\ref{thm:transition}, $\rho_n^2 \to 1-\tau^{-2}>0$ whenever $\tau>1$. Since $\Mat_1(\vu) = \vv {\vv^{\boxtimes(q-1)}}\T$, we consequently have
\begin{equation}\label{eq:recursive-decomposition}
\Mat_1(\hat\vu) = \rho_n \, \vv {\vv^{\boxtimes(q-1)}}\T + \sqrt{1-\rho_n^2} \, \mG.
\end{equation}
The first term in~\eqref{eq:recursive-decomposition} is rank one with singular value $\rho_n$, bounded away from zero. Thus, if one could establish
\begin{equation}\label{eq:recursive-remainder}
\|\mG\|_{\mathrm{op}} = o(1),
\end{equation}
then the Davis--Kahan/Wedin perturbation bound would imply $|\hat\vv^\top\vv|\to 1$.

In the asymmetric model, \citet{montanari_statistical_2014} establish the analogue of~\eqref{eq:recursive-remainder} in the proof of their Theorem~5, with the sharper bound $\|\mG'\|_{\mathrm{op}}=O(n^{-1/2})$, almost surely. We expect the same estimate to hold in the symmetric model, with a negligibility argument similar to Lemma~\ref{lem:permut}. Establishing~\eqref{eq:recursive-remainder} rigorously would, however, require an operator-norm estimate for the matricized eigenvector remainder that is not provided by our present deterministic-equivalent analysis. We therefore leave this step as a conjecture.
\end{proof}
The numerical experiments in Figure~\ref{fig:recursive} support Conjecture~\ref{conj:recursive}.

\section{Choice of Unfolding}\label{sec:choice_q}
We saw in Section~\ref{sec:discussion} that, if the dependencies induced by the high-order symmetries are spectrally negligible for the $n_k \times n_{p-k}$ symmetric unfolding, then the disproportionate random matrix results of~\citet[Theorem~1]{feldman_spiked_2023} suggest the following sharp phase transition.
\begin{conjecture}[Phase Transition]\label{conj:transition}
Let $\tX = \lambda \, {\vv}^{\otimes p} + \tZ$, with $\lambda > 0$, $\vv \in\mbS^{n-1}$ and $\tZ \sim \GOTE_{p, n}$. Let $k < p/2$, $\mX \coloneqq \Phi_k(\tX) \in \mbR^{n_k \times n_{p-k}}$, and let $(\hat\vv_{\mathrm{\ell}}, \hat\vv_{\mathrm{r}}) \in \mbS^{n_k-1}\times\mbS^{n_{p-k}-1}$ denote its leading left and right singular vectors. Writing $\tau_{\mathrm{\ell}} \coloneqq {\lambda} / {(\gamma_{\mathrm{\ell}, p, k} \, n^{(p-2)/4})}$, with $\gamma_{\mathrm{\ell}, p, k} \coloneqq ({k!(p-k)!} / {(p-1)!^2})^{1/4}$ and $\tau_{\mathrm{r}} \coloneqq {\lambda} / {(\gamma_{\mathrm{r}, p, k} \, n^{(p-k-1)/2})}$, with $\gamma_{\mathrm{r}, p, k} \coloneqq (k! / (p-1)!)^{1/2}$, one has
\begin{equation}\label{eq:phase_transition_conj}
    \big|{\hat\vv_{\mathrm{\ell}}}^{\T}\vv^{\odot k}\big|^2 \aslim \big(1-\tau_{\mathrm{\ell}}^{-4}\big)^+ \qquad \text{and} \qquad 
    \big|{\hat\vv_{\mathrm{r}}}^{\T}\vv^{\odot p-k}\big|^2 \aslim \frac{\tau_{\mathrm{r}}^2}{1 + \tau_{\mathrm{r}}^2}.
\end{equation}
\end{conjecture}
This conjecture is conditional on a negligibility result analogous to Lemma~\ref{lem:permut}, showing that the contractions associated with indices $(\vi',\vk') \in \mathcal I_k \times \mathcal I_{p-k}$ such that $\vi'\sqcup\vk'=\vi\sqcup\vk$ are spectrally negligible. 
By studying the function $k \mapsto k!(p-k)!$, we see that the sharp constant in the phase-transition threshold for the short-side singular vector is minimized by an ``as square as possible'' unfolding. This is consistent with the recommendation of~\citet[Lemma~3.1]{montanari_statistical_2014}, while highlighting an important distinction with the asymmetric-noise setting of~\citet{arous_long_2023}. More precisely, the result of~\citet[Lemma~3.1]{montanari_statistical_2014} concerns partial recovery on the \emph{long side} of the unfolding, for which the relevant threshold has a larger order of magnitude in the disproportionate regime, namely $n^{(p-k-1)/2}$. Thus, their recommendation should be interpreted with some caution in light of the distinction between the two singular-vector sides. Nevertheless, our results show that, for the \emph{short-side} singular vector, their ``as square as possible'' prescription is indeed optimal at the level of the sharp constant (although not at the level of the polynomial order, which is the same for all unfoldings, namely $n^{(p-2)/4}$).

\begin{figure}
    \centering
    \includegraphics[width=0.8\linewidth]{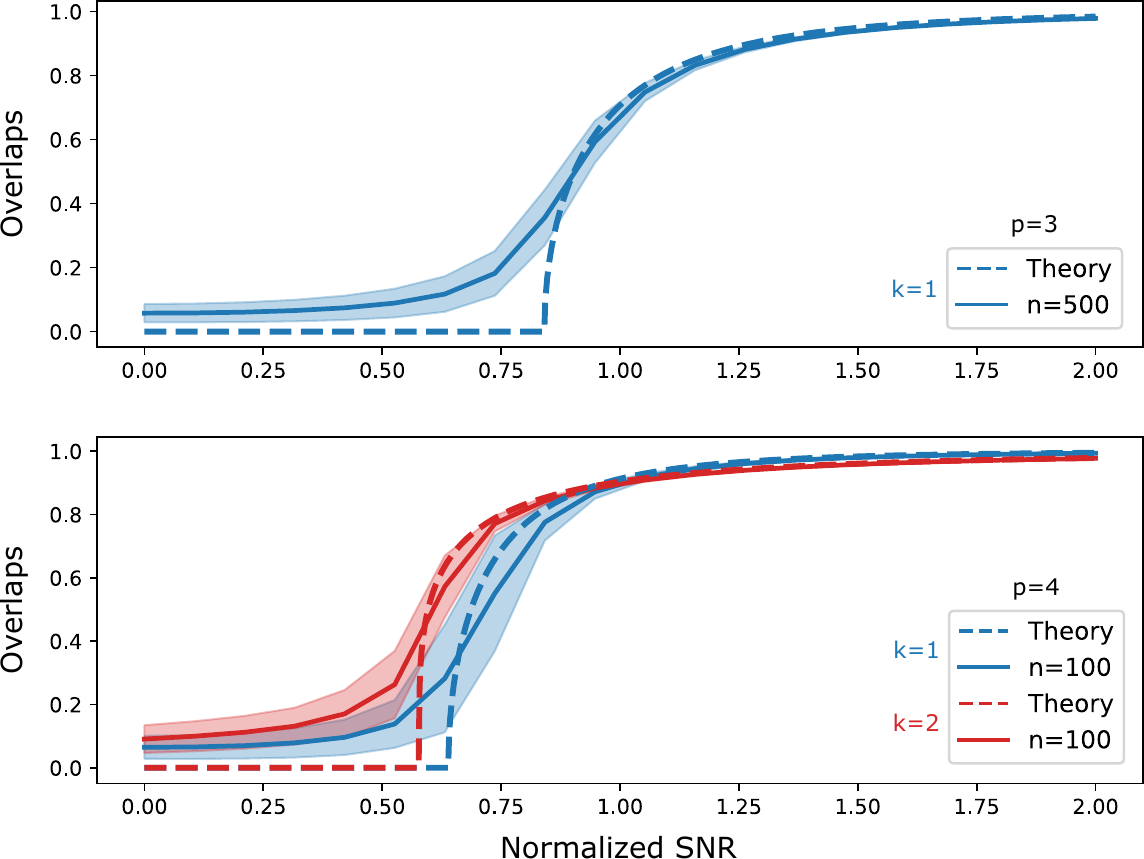}
    \caption{Theoretical and simulated eigenvalues and overlaps for the symmetrically unfolded spiked tensor $\mM \coloneqq \Phi_k(\lambda \, {\vv}^{\otimes p} + \tZ)$, with $\tZ \sim \GOTE_{p, n}$, for $(p,k,n) = (3,1,500)$ (top) and $(p,k,n) = (4,\{1, 2\},100)$ (bottom). See Conjecture~\ref{conj:transition}.}
    \label{fig:choice_q}
\end{figure}

\begin{figure}
    \centering
    \includegraphics[width=0.8\linewidth]{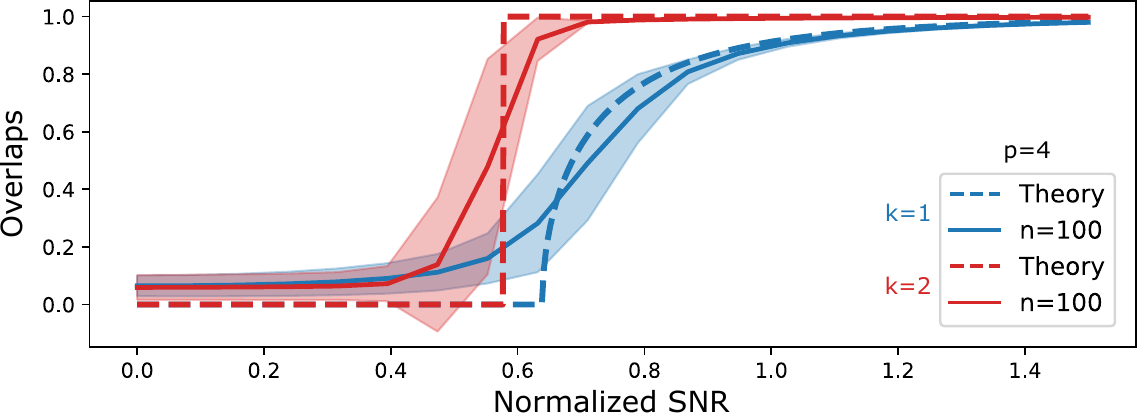}
    \caption{Theoretical and simulated overlaps for the recursively symmetrically unfolded spiked tensor $\mM \coloneqq \Phi_k(\lambda \, {\vv}^{\otimes p} + \tZ)$, with $\tZ \sim \GOTE_{p, n}$, for $p=4$, $k \in \{1, 2\}$ and $n=100$ (see Conjectures~\ref{conj:recursive} and~\ref{conj:transition}).}
    \label{fig:choice_recursive}
\end{figure}

We compare this conjecture with numerical simulations in Figure~\ref{fig:choice_q}, for $(p,k,n) = (3,1,500)$ and $(p,k,n) = (4,\{1, 2\},100)$. The simulations match the conjectured overlaps. For $p=4$, the ``as square as possible'' unfolding ($k=2$) undergoes a phase transition before the ``as large as possible'' unfolding ($k=1$), resulting in better partial recovery. This superiority is reinforced with recursive unfolding; see Figure~\ref{fig:choice_recursive} for numerical simulations.

From the computational standpoint, the comparison is somewhat different. The square unfolding involves extracting the leading $r$ eigenpairs of a symmetric $n_{p/2}\times n_{p/2}$ matrix, which costs $O(r n_{p/2}^2) = O({r n^p}/{(p/2)!^2})$ using an iterative eigensolver---up to the number of iterations required for convergence. A non-square unfolding involves extracting the leading $r$ left singular vectors of a $n_k\times n_{p-k}$ matrix. An iterative SVD requires $O(r n_k n_{p-k})$ arithmetic operations per iteration, giving $O({r n^p} / {(k!(p-k)!)})$ up to the same iteration-count considerations. Thus, in principle both types of unfolding have the same polynomial computational complexity, but the leading combinatorial factor is minimized by the ``as large as possible'' unfolding ($k=1$). Moreover, when $k=1$, the left singular vectors directly provide an estimate of $\vv\in\mbS^{n-1}$, without requiring a subsequent recursive unfolding step.

Overall, this suggests a tradeoff between statistical and computational efficiency. From a recovery standpoint, an ``as square as possible'' unfolding ($k=\lfloor p/2\rfloor$) is preferable because it minimizes the conjectured phase-transition threshold, while recursive unfolding can then be used to recover the planted signal $\vv$. From a computational standpoint, an ``as large as possible'' unfolding ($k=1$) is preferable: it has the smallest leading combinatorial factor and directly produces an estimator in the original signal space. To the best of our knowledge, this is the first analysis of the choice of unfolding to exhibit such a tradeoff between computational efficiency and estimation performance. It highlights a qualitative distinction between symmetric and asymmetric tensor noise: in the asymmetric setting, the recovery threshold is independent of the choice of unfolding~\citep{arous_long_2023,feldman_sharp_2023}. It is also reminiscent of the hierarchy of statistical--computational tradeoffs provided by Kikuchi matrices~\citep{wein_kikuchi_2025,bandeira_matrix_2026}.

\vskip 0.2in
\bibliography{sample}

\end{document}